\documentclass[conference]{IEEEtran}
\IEEEoverridecommandlockouts
\usepackage{cite}
\usepackage{multirow}
\usepackage{tabu}
\usepackage{tabularx}
\usepackage[pdftex]{graphicx}
\graphicspath{{../pdf/}{../jpeg/}}
\DeclareGraphicsExtensions{.pdf,.jpeg,.png}
\usepackage[cmex10]{amsmath}
\usepackage{amssymb}
\usepackage{algorithmic}
\usepackage{array}
\usepackage{mdwmath}
\usepackage{eqparbox}
\usepackage{caption}
\usepackage{subcaption}
\usepackage{url}
\usepackage{booktabs}
\usepackage{color}
\usepackage{tikz-network}
\usepackage{bm}
\usepackage[linesnumbered,ruled]{algorithm2e}
\usepackage{flushend}
\usepackage{csquotes}
\usepackage{url}

\newcommand\T{\rule{0pt}{2.9ex}}       
\newcommand\B{\rule[-1.2ex]{0pt}{0pt}} 

\begin{document}

\title{Not Half Bad: Exploring Half-Precision in Graph Convolutional Neural Networks}

\author{
\IEEEauthorblockN{John Brennan \IEEEauthorrefmark{1}\IEEEauthorrefmark{2},
Stephen Bonner \IEEEauthorrefmark{1},
Amir Atapour-Abarghouei \IEEEauthorrefmark{1},\\
Philip T Jackson \IEEEauthorrefmark{1},
Boguslaw Obara \IEEEauthorrefmark{2} and
Andrew Stephen McGough \IEEEauthorrefmark{1}}

\IEEEauthorblockA{\IEEEauthorrefmark{1}School of Computing, Newcastle University, Newcastle, UK \\ \{john.brennan, stephen.bonner3, amir.atapour-abarghouei, philip.jackson, stephen.mcgough\}@newcastle.ac.uk}
\IEEEauthorblockA{\IEEEauthorrefmark{2}Department of Computer Science, Durham University, Durham, UK \\ \{j.d.brennan, boguslaw.obara\}@durham.ac.uk}
}
\maketitle

\begin{abstract}

    With the growing significance of graphs as an effective representation of data in numerous applications, efficient graph analysis using modern machine learning is receiving a growing level of attention. Deep learning approaches often operate over the entire adjacency matrix -- as the input and intermediate network layers are all designed in proportion to the size of the adjacency matrix -- leading to intensive computation and large memory requirements as the graph size increases. It is therefore desirable to identify efficient measures to reduce both run-time and memory requirements allowing for the analysis of the largest graphs possible. The use of reduced precision operations within the forward and backward passes of a deep neural network along with novel specialised hardware in modern GPUs can offer promising avenues towards efficiency. In this paper, we provide an in-depth exploration of the use of reduced-precision operations, easily integrable into the highly popular \emph{PyTorch} framework, and an analysis of the effects of Tensor Cores on graph convolutional neural networks. We perform an extensive experimental evaluation of three GPU architectures and two widely-used graph analysis tasks (vertex classification and link prediction) using well-known benchmark and synthetically generated datasets. Thus allowing us to make important observations on the effects of reduced-precision operations and Tensor Cores on computational and memory usage of graph convolutional neural networks -- often neglected in the literature.

\end{abstract}


\section{Introduction}
\label{sec:introduction}

Graphs, which represent a number of entities (referred to as Vertices) and the links between these entities (referred to as Edges), have become an indispensable tool for analysis of data across many disciplines including social sciences, security and medicine. We define a graph $G = (V,E)$ as a set of vertices $V$, with a corresponding set of edges $E$. $E$ is composed of unordered tuples ${u,v}$ where $u, v \in V$. Their ability to represent the links between different entities makes them a more natural representation for tasks such as identifying relationships between different entities (link prediction) than other data representation formats.

As in many other fields, deep learning is helping to revolutionise the area of graph analytics. Historically, graphs have been analysed through kernel-based methods~\cite{Kriege2020ASO}, however, recent advances in the area of Graph Convolutional Networks (GCN) have shown great promise for improved results. Although other approaches towards graph analysis have subsequently been developed, these have yet to diminish the need for GCNs~\cite{shchur2018pitfalls}. A GCN layer is a learnable non-linear function of the vertex features from the previous layer (represented as a matrix) and the adjacency matrix\footnote{An adjacency matrix $A$ is an $n$ by $n$ matrix, where $n$ is the number of vertices in the graph. $[a,b]=1$ indicates an edge between vertices $a$ and $b$.} for the graph. As such, GCNs are almost entirely constructed from matrix operations and hence well amenable to GPU programming.

One of the main drawbacks to the use of GCNs is the memory footprint. Unlike other forms of Deep Learning where only a subset of the data ever needs to be processed at any given point in time, a GCN operates on the entire adjacency matrix -- i.e. the entire graph -- at each step. This limits the size of the graph which may be operated upon. This is compounded by the fact that most GCN functions comprise of a number of matrices of commensurate size to the adjacency matrix.

The use of reduced-precision computation, most often combining half and full precision floating-point values, has been demonstrated to be beneficial within other areas of deep learning, significantly reducing memory requirements and training time and improving performance \cite{micikevicius2017mixed, das2018mixed, kuchaiev2018mixed}. As memory requirements are the biggest limitation to GCNs, applying reduced-precision computations would seem an obvious line of attack in order to process larger and more complex graphs.

Recent advances in GPU technology have lead to the introduction of Tensor Cores, which enable dynamic adaptation of mixed-precision floating-point computations. These allow for acceleration of Deep Learning and, according to NVIDIA, can provide up to ten times speed up~\cite{tensorcores}. This, again, could be most useful when training larger GCNs.

In this work, we implement four levels of optimisation, with respect to the operation precision levels used in the forward and backward passes of the network, on two types of GCNs -- a standard GCN and a Graph Convolutional Auto-Encoder (GAE) -- in order to evaluate the advantages and disadvantages of each optimisation level. Our experiments range from using no optimisation (full precision for all operation) through to everything performed and stored in half-precision. In order to make use of the Tensor Cores, certain model parameters, including input size, need to be divisible by 8. We therefore evaluate results both with and without padding the data to be a multiple of 8. 

Based on prior literature which used reduced precision in deep learning, such as computer vision and natural language processing, one would Na\"ively assume that learning-based graph analysis tasks would see reduced memory requirements, faster training times and perhaps even improved performance -- with the desire that memory reduction would be the most significant. As such, we set these as our hypotheses for this work and design the experiments accordingly to evaluate the validity of each point. 
We use the real-world Cora\footnote{\url{https://relational.fit.cvut.cz/dataset/CORA}} dataset and to allow the scaling of graphs to specific sizes, we also generate synthetic graphs using the Barab\'{a}si-Albert Model \cite{Barabasi2000}.

In short, this paper attempts to answer the following important questions with respect to the effects of reduced-precision operations and Tensor Cores on graph-based neural networks:

\begin{itemize}

    \item \emph{Run-Time} - Will using reduced-precision operations and Tensor Cores lead to more efficient training time for graph convolutional neural networks?

    \item \emph{Memory Usage} - Will using reduced-precision operations and Tensor Cores lead to reduced memory requirement for graph convolutional neural networks?

    \item \emph{Predictive Performance} - Will using reduced-precision operations and Tensor Cores lead to an improvement, degradation or no significant change in the predictive performance of graph convolutional neural networks?

\end{itemize}

To the best of our knowledge, this is the first work to conduct a comprehensive analysis of the effects of reduced precision on graph convolutional neural networks. The code is available at \url{https://github.com/grossular/half-precision-gcn}.

{\color{red}
















}


\section{Background and Motivation}
\label{sec:background}

The introduction of Tensor Cores, in the NVIDIA Volta GPU microarchitecture~\cite{volta}, improved deep learning performance in comparison to conventional CUDA cores. Tensor Cores enable mixed-precision computing (combining features from both half and full-precision operations), dynamically adapting calculations to accelerate throughput while preserving the accuracy. It is claimed that this technology can provide up to 10$\times$ training speed up across a variety of workloads \cite{tensorcores}.


To take advantage of Tensor Cores workloads must use \emph{mixed-precision} computations. Deep learning traditionally uses full-precision (\texttt{FP32}), however, demands from ever larger datasets and thus network architectures leads to longer training times and higher memory requirements in order for the neural network to converge. If indeed this is possible. Half-precision (e.g. \texttt{FP16}) can address memory and computation issues. However, \texttt{FP16} has a narrower dynamic range than \texttt{FP32} leading to significant loss of accuracy -- potentially preventing successful training. Mixed-precision operations are used to address this -- maintaining accuracy standards while taking advantage of reduced computation and memory bandwidth.


Mixed-precision operations are not the only requirement for using Tensor Cores -- certain layer parameters (e.g. batch size, input size, output size, number of channels) also need to be \emph{divisible by 8}. A requirement on how data is stored and accessed in memory \cite{performanceGuide}. Tensor Cores primarily optimise GEMM (General Matrix Multiplications), a fundamental building block for many operations in neural networks, such as fully-connected layers, recurrent layers and convolutional layers. Mixed-precision use of Tensor Cores has been extensively investigated for various tasks such as computer vision and natural language processing \cite{micikevicius2017mixed, kuchaiev2018mixed}. However, the important area of graph analysis has thus far been neglected.


We investigate the applicability of Automatic Mixed Precision (AMP) on two commonly-used graph network architectures (GCN and GAE). NVIDIA Apex \cite{apex} enables AMP via PyTorch \cite{pytorch} -- providing the primary framework for our experiments. Apex AMP provides the opportunity to easily experiment with different levels of precision, by selecting an \enquote{optimisation level}. Four default AMP modes (optimisation levels) are provided, briefly described in the following:


\textbf{O0:} Enables full-precision \texttt{FP32} training. Neural network weights and their corresponding operations are \texttt{FP32}. As this makes no modifications it can be used to establish a baseline.


\textbf{O1:} The most commonly-used AMP mode, places each operation into one of two lists: a whitelist for all \textit{Tensor Core-friendly} operations (e.g. GEMM and convolutions), and a blacklist for all others (e.g. non-linear operations, normalisations). Whitelist operations are performed in \texttt{FP16} and blacklist in \texttt{FP32}. Dynamic loss scaling is also important, as activation gradient values during \texttt{FP16} training need to be scaled to preserve values that could otherwise be lost to zero.


\textbf{O2:} (\enquote{Almost-\texttt{FP16} Mixed Precision}), casts the model weights to \texttt{FP16} but maintains a set of \texttt{FP32} master weights. The input data being fed through the network is cast to \texttt{FP16} but the optimiser acts directly on the \texttt{FP32} weights. Dynamic loss scaling is implemented as in O1. O1 and O2 are essentially different implementations of mixed precision and their performance will depend on the type of data and the operations involved in the network architecture.


\textbf{O3:} Enables full \texttt{FP16} across all operations and as such does not achieve the stability of O1 and O2 and will lead to a loss of accuracy. Similar to O0, this mode is primarily used to provide a baseline for the evaluation of other levels.


Our experiments will enable an accurate and detailed analysis of the effects of mixed-precision on GCNs in terms of model performance, memory footprint and run-time.


\section{Related Work}
\label{sec:litreview}

With the growing popularity of deep neural networks and hence the increasing focus on training and inference efficiency, the use of reduced precision has received significant attention within the existing literature \cite{ginsburg2017training, micikevicius2017mixed, gupta2015deep, hubara2017quantized, courbariaux2014training, wang2018training}. For instance, there have been attempts to binarise model weights and activations while gradient calculation is kept within the full-precision format \cite{hubara2017quantized}. In \cite{rastegari2016xnor}, even gradients are binarised along with all other tensors in order to improve training and inference efficiency in terms of both memory usage and run-time. However, despite their impressive computational efficiency, such approaches always lead to significant losses in accuracy with larger model architectures.

To resolve the issue of accuracy loss, the majority of the recent work has shifted towards using at least 16 bits for data and gradient computation. The approach proposed in \cite{micikevicius2017mixed} uses a 16-bit floating-point format accumulating results into 32-bit arrays and ensures gradients with a small magnitude are preserved via loss-scaling. Accuracy is also maintained in \cite{das2018mixed, koster2017flexpoint} using a custom format with a 16-bit mantissa and a shared exponent to train large neural networks. Despite their promising performance, such approaches keep a 32-bit copy of the model weights to enable precise weight updates and partial products are accumulated in a 32-bit format.

In \cite{wang2018training}, 8-bit floating-point numbers are used for both the numerical representation of the data and all the computations required for the operations involved in the forward and backward passes of the model. The approach outlined in \cite{mellempudi2019mixed} enhances the use of 8-bit floating-point representation by compensating for the reduced subnormal range of 8-bit floating-point representation for improved error propagation leading to better model accuracy.

It is important to note, however, that the use of any reduced-precision approach, such as those reviewed above, heavily depends on model size, input data modality and the nature of the task. Extensive exploration and benchmarking of various reduced-precision methodologies have been carried out for Convolutional Neural Networks and Transformer models for computer vision and natural language processing tasks \cite{micikevicius2017mixed, kuchaiev2018mixed}, whilst the use of neural networks for graph-based applications is not yet fully investigated. Consequently, in this paper, we attempt to provide a detailed study of the use of mixed-precision operations using specialised hardware for graph convolutional neural networks.

\section{Methodology}
\label{sec:method}

The primary objective of this paper is to investigate how graph-specific neural models are affected with the use of reduced-precision operations. In order to achieve this, experiments are run on real and synthetic graphs, over all available optimisation levels, on hardware equipped both with and without Tensor Cores. In the remainder of this section, we give a brief overview of the neural network architectures used in this work, before detailing the changes made for this study.

\subsection{Graph Convolutions}
\label{sec:gcn}

 Here, we introduce the basics of Graph Convolutional Networks (GCN) \cite{kipf2017semi} and detail how they may be affected by the move to reduced precision. GCNs can be thought of differentiable functions for aggregating feature representations from the neighbourhood of a given vertex \cite{hamilton2017inductive}. For initial input, a GCN-based model takes the normalised adjacency matrix $\hat{\mathbf{A}}$ representing a graph $G$, and a matrix of initial vertex level features $\mathbf{X}$, and computes a new matrix of vertex level features $\mathbf{H} = GCN(\hat{\mathbf{A}}, \mathbf{X})$. Whilst $\mathbf{X}$ can be initialized with pre-computed vertex features, it is common to initialize it with one-hot feature vectors when no prior knowledge is available (in which case $\mathbf{X}$ is the identity matrix $\mathbf{I}$). Each layer in a GCN performs the following operation \cite{kipf2017semi}:
\begin{equation}
    \label{eq:GCN}
    GCN^{(l)}(\mathbf{H}^{(l)}, \hat{\mathbf{A}}) = \sigma_r (\hat{\mathbf{A}} \mathbf{H}^{(l-1)} \mathbf{W}^{(l)}) \, ,
\end{equation}
where $l$ is the number of the current layer, $\mathbf{W}^{(l)}$ denotes the weight matrix of that layer, and $H^{(l-1)}$ refers to the features computed at the previous layer or is equal to $\mathbf{X}$ at $l=0$.

A GCN function can be considered as performing a weighted average of the neighbourhood features for each vertex in the graph. Stacking multiple GCN layers has the effect of increasing the number of hops from which a vertex-level representation can aggregate information -- a three-layer GCN will aggregate information from three-hops within the graph to create each representation.

One interesting thing to consider is the dimensionality of the matrices involved in the GCN operation in Equation \ref{eq:GCN}:

\begin{itemize}
    \item The adjacency matrix, $\mathbf{A}$, is of size $N_v \times N_v$, where $N_v = |V|$ is the number of vertices in the graph.
    \item The input features matrix, $\mathbf{X}$, is of size $N_v \times F_v$, where $F_v$ is the number of features for each vertex. Where no vertex features are present and the identity matrix $\mathbf{I}$ is used, the dimensionality would again be $N_v \times N_v$.
    \item The parameter matrix, $\mathbf{W}$, is of size $F_v \times d$, where $d$ is the number of units in that layer.
\end{itemize}

One thing to note is that the number of model parameters is closely tied to the size of the input features and that the resulting output from each layer in a GCN is bound by the number of vertices, in contrast to computer vision models.

\subsection{Graph Convolutional Auto-Encoders}
\label{sec:gae}

GCNs are trained via supervised learning, where labels are provided for a specific task -- commonly vertex classification \cite{hamilton2017inductive, kipf2017semi}. However, extensions have been made to allow for convolutional auto-encoders for graph datasets, called Graph Auto Encoders (GAE) \cite{kipf2016variational}. Auto-encoders are a type of un-supervised neural network which compressed the input data to a low-dimensional space, and then reconstructs the original data from the learned representation. This is commonly performed in order to use the resulting embeddings for the task of link prediction \cite{kipf2016variational, bonner2019temporal, bonner2018temporal}.

Here, we consider a non-probabilistic version of the GAE, where the goal is to learn a low-dimensional representation of $\mathbf{A}$ from $G$, via an encoding from a GCN $\mathbf{Z} = GCN(\mathbf{A}, \mathbf{X})$, such that it can be used to accurately reconstruct the graph via a product between $\mathbf{Z}$ and its transpose passed through an element-wise logistic function $\sigma$:

\begin{equation}
    \label{eq:GAE}
    \mathbf{A}^{\prime} = \sigma(\mathbf{Z} \mathbf{Z}^\mathsf{T}).
\end{equation}

\begin{figure}[b]
  \centering
    \begin{subfigure}[b]{0.24\textwidth}
      \includegraphics[width=\textwidth]{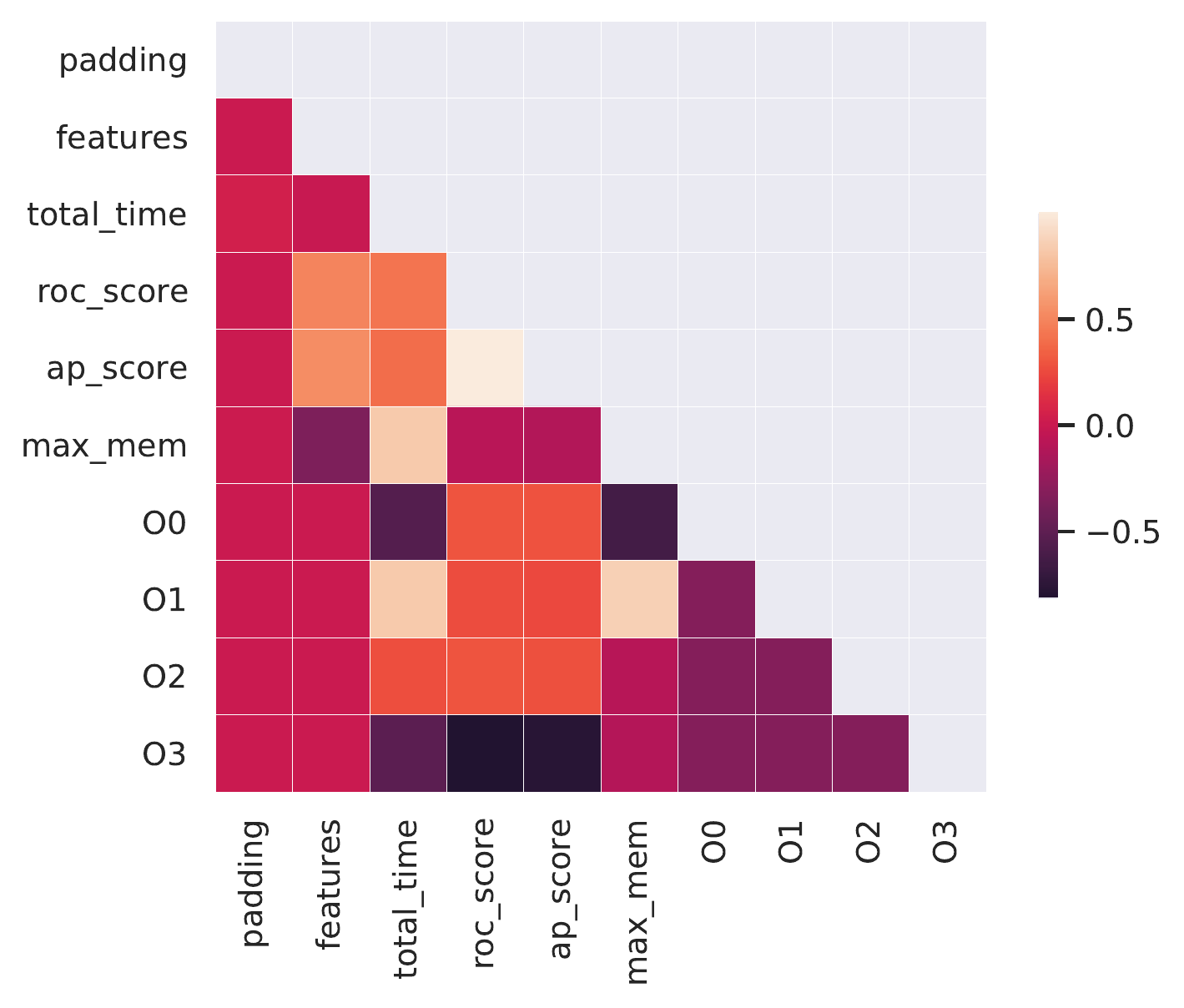}
      \caption{GAE}
    \end{subfigure}
    \begin{subfigure}[b]{0.24\textwidth}
      \includegraphics[width=\textwidth]{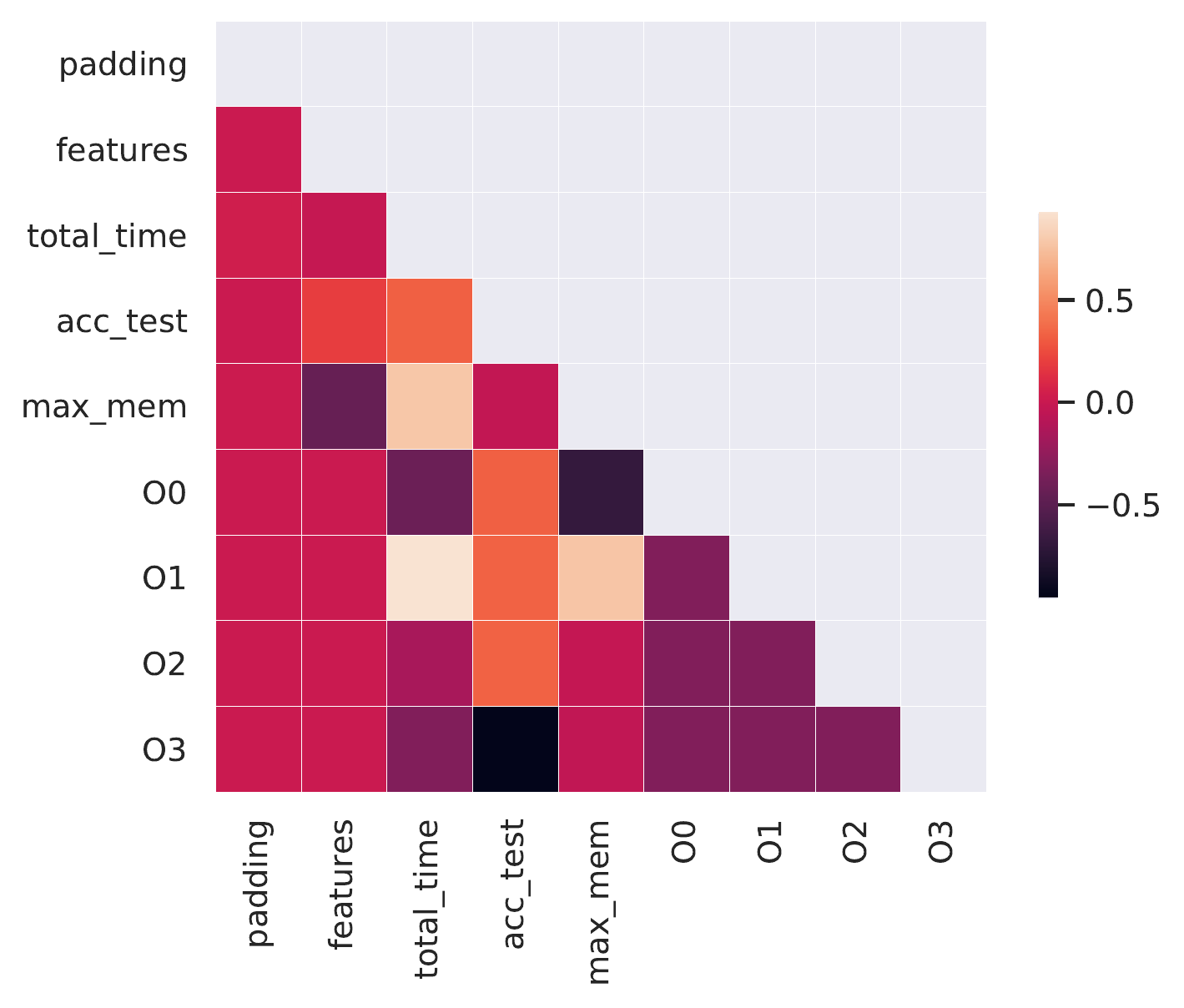}
      \caption{GCN}
    \end{subfigure}
    \caption{Correlation of predictive performance values on the Cora dataset for the GCN and GAE models using the V100.}
    \label{fig:acc-corr}
    \vskip -10pt
\end{figure}

\begin{figure}[b]
  \centering
    \begin{subfigure}[b]{0.24\textwidth}
      \includegraphics[width=\textwidth]{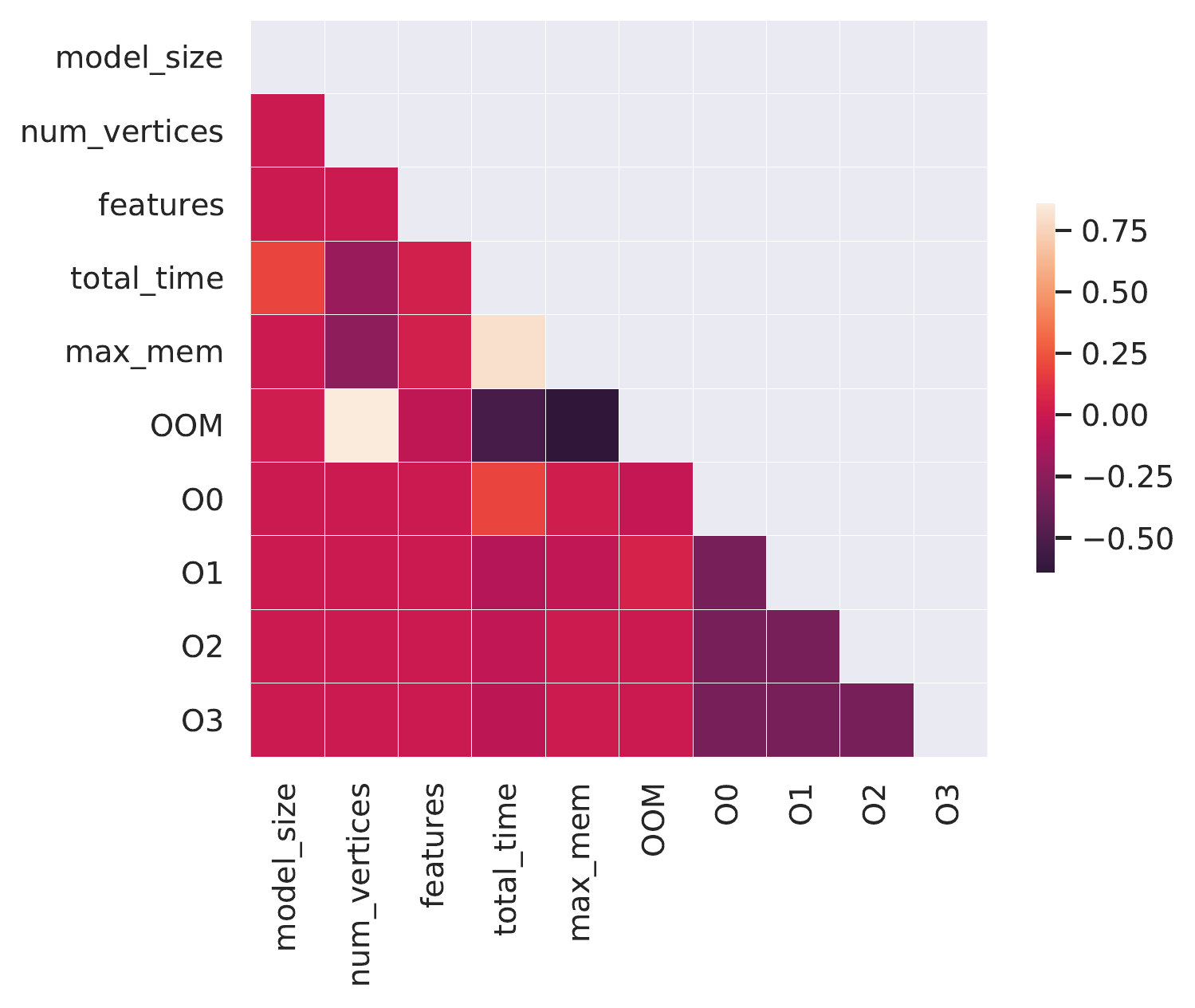}
      \caption{GAE}
    \end{subfigure}
    \begin{subfigure}[b]{0.24\textwidth}
      \includegraphics[width=\textwidth]{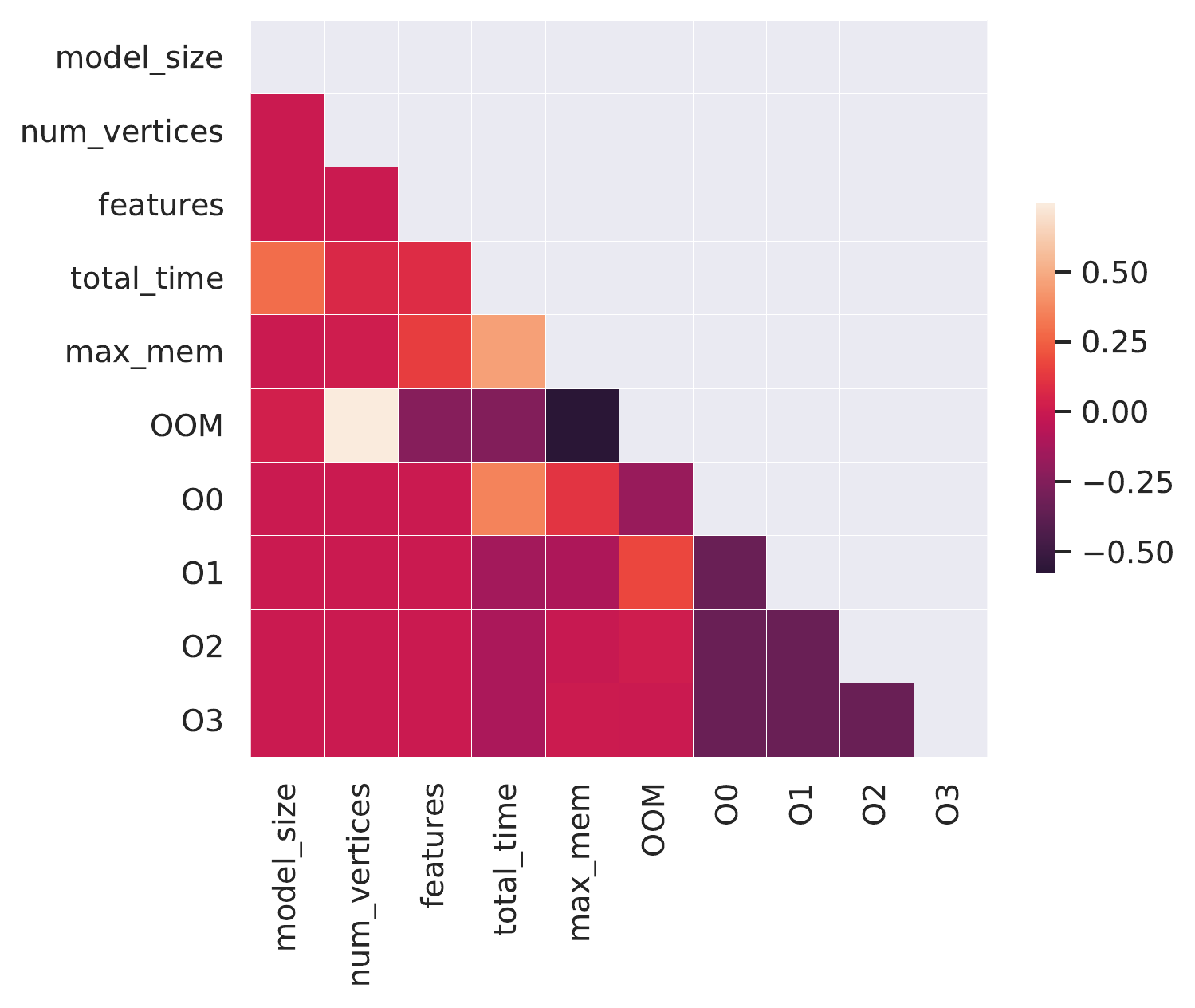}
      \caption{GCN}
    \end{subfigure}
    \caption{Correlation of run-time experiment on the GCN and GAE models using the V100.}
    \label{fig:mem-corr}
    \vskip -10pt
\end{figure}

\begin{figure*}[t!]
  \centering
    \begin{subfigure}[b]{0.28\textwidth}
      \includegraphics[width=\textwidth]{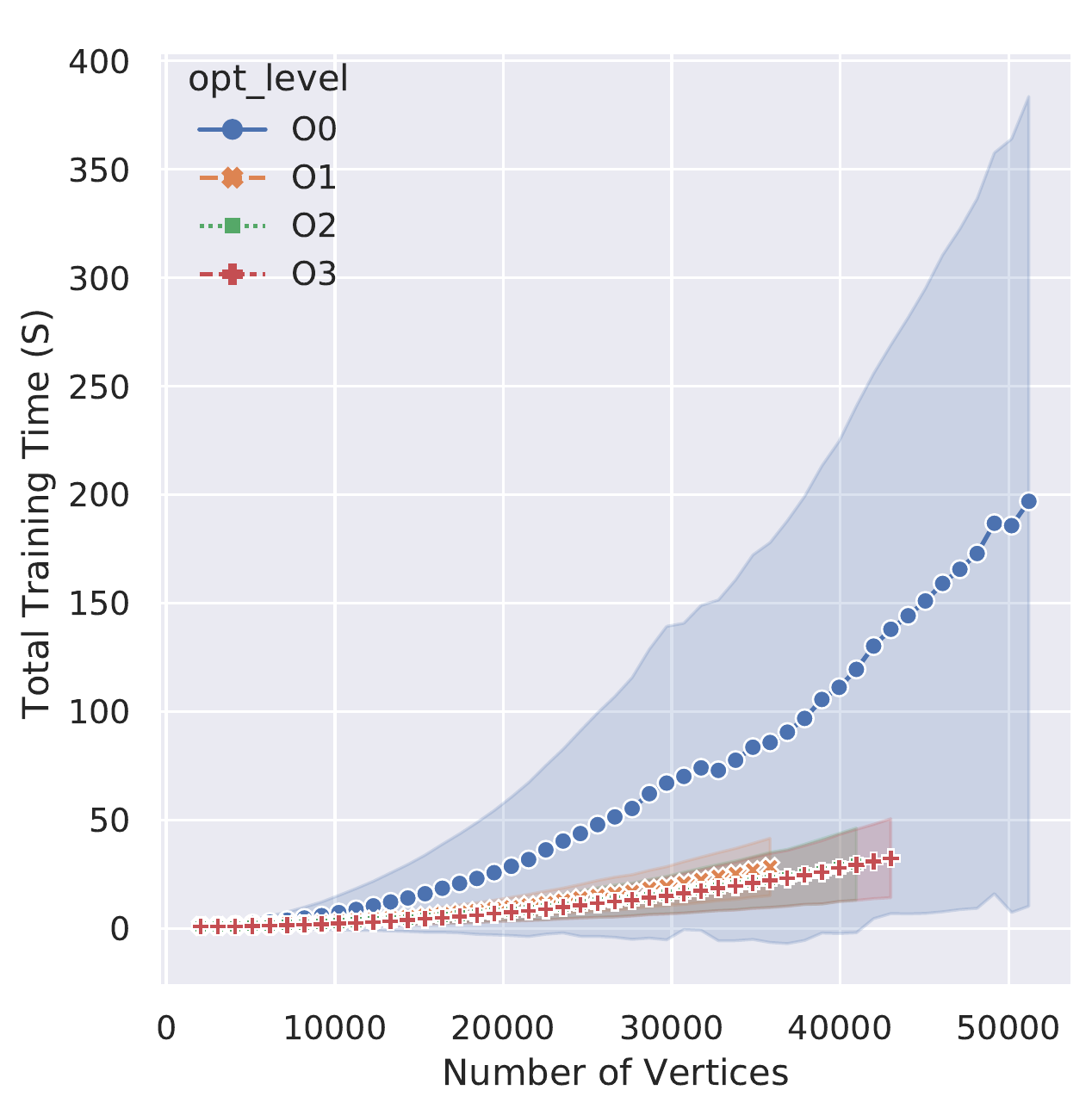}
      \caption{Titan RTX}
      \label{fig:gcn-vertex-runtime-rtx}
    \end{subfigure}
    \begin{subfigure}[b]{0.28\textwidth}
      \includegraphics[width=\textwidth]{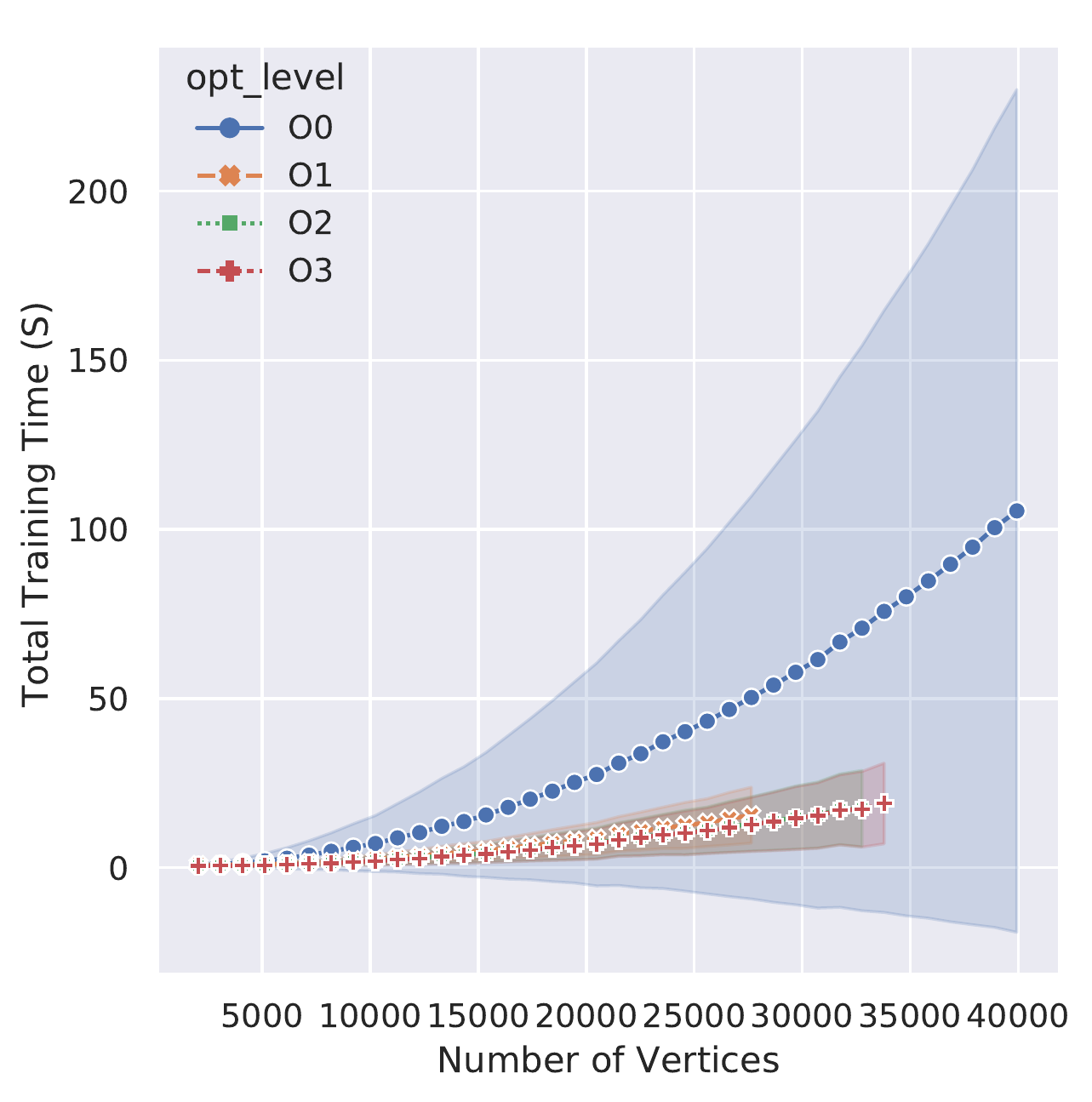}
      \caption{V100}
      \label{fig:gcn-vertex-runtime-v100}
    \end{subfigure}
    \begin{subfigure}[b]{0.28\textwidth}
      \includegraphics[width=\textwidth]{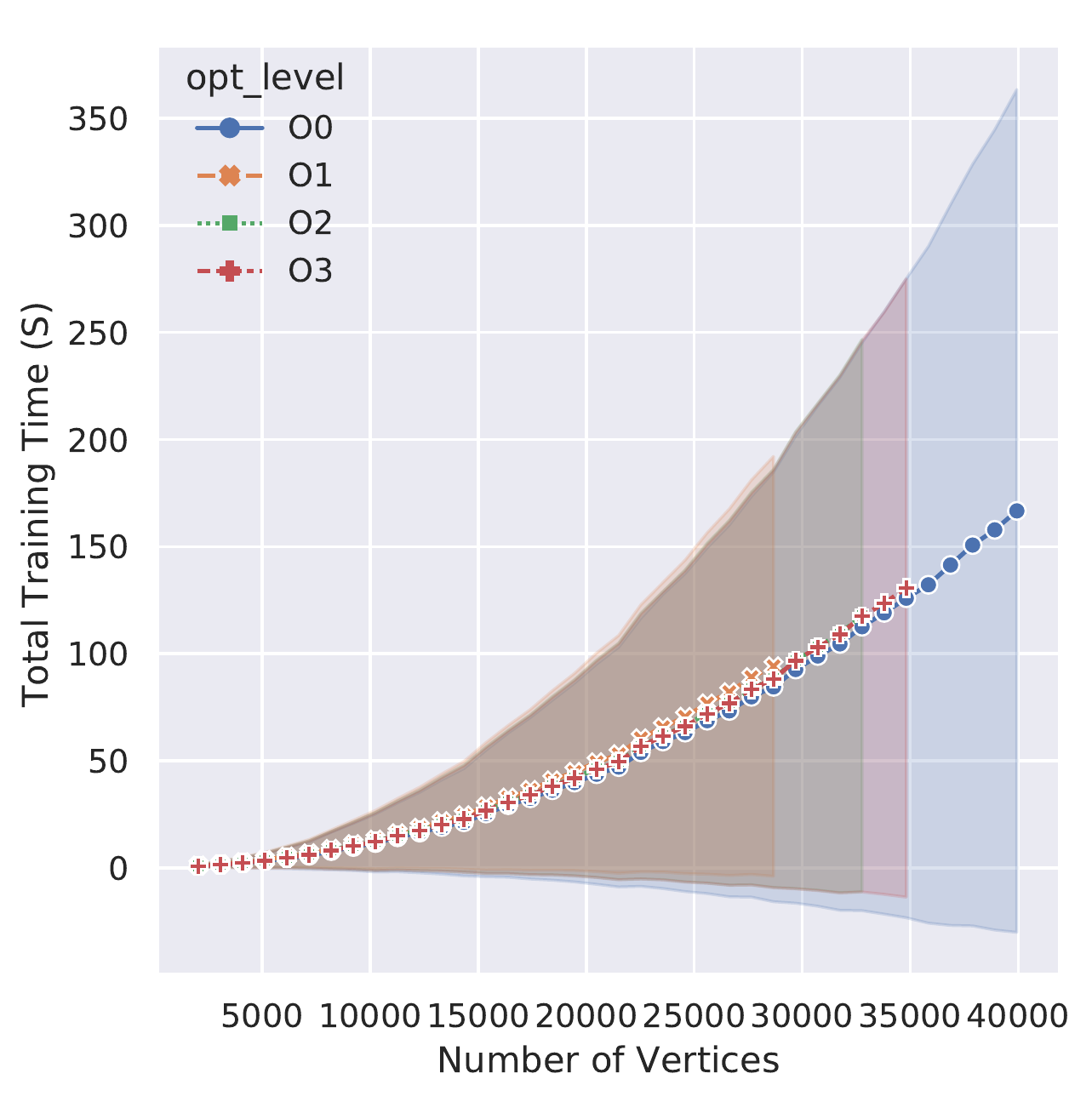}
      \caption{P100}
      \label{fig:gcn-vertex-runtime-p100}
    \end{subfigure}
    \caption{GCN total training time versus increases in graph size.}
    \label{fig:gcn-vertex-runtime}
    \vskip -10pt
\end{figure*}

\subsection{Reduced Precision Changes}

For the models used in this paper, we replicated the overall architecture from the prior works \cite{kipf2017semi,kipf2016variational}. However, some changes were required to ensure suitability for processing using reduced precision. In order to take advantage of the Tensor Cores, available on Volta and subsequent NVIDIA GPU architectures, some padding of the input graph is required under certain conditions. This is because Tensor Cores are only activated when specific matrices involved in operations for the forward and backward passed are divisible by 8, for \texttt{FP16} matrices, or 16, for \texttt{INT8} matrices \cite{performanceGuide}. Due to this, in our experiments, where an input matrix is observed to be indivisible by 8 it is padded with zeroes to make it adhere to this condition so that Tensor Cores could be fully utilised. This padding can be though of as additional vertices added to the graph with no edges to it's self of other vertices. These added vertices are removed before the loss commutation is performed. We present experimental evaluation of this in Section \ref{sec:results} to assess if this process has any negative impact of predictive performance. Additionally, due to current limitations in PyTorch\footnote{https://github.com/pytorch/pytorch/issues/41069}, all tensors needed to be cast to dense matrices in order to be able to use reduced precision modes.

\section{Experimental Setup}
\label{sec:experiment_setup}

\subsection{Datasets}

Our experiments use two primary datasets: the real world benchmark Cora dataset \cite{yang2016revisiting} and a synthetic dataset using the well-known Barab\'{a}si-Albert Model graph generation model \cite{Barabasi2000}. The Cora dataset has been used for vertex classification and link prediction in the original GCN \cite{kipf2017semi} and GAE \cite{kipf2016variational} papers, so was the ideal choice for assessing any predictive performance changes due to the reduced precision. For observing run time and memory related metrics, synthetics Barab\'{a}si-Albert graphs where used as they reflect the scale-free nature of many empirical graphs and allowed us to precisely control the number of vertices in the input graph \cite{Barabasi2000}.

\subsection{Experimental Environment}

The performance of the models was measured on three different compute systems, with three different generations of NVIDIA GPU (Pascal, Volta and Turing). The V100 (Volta) and Titan RTX (Turing) both are equipped with Tensor cores, whilst the P100 (Pascal) has no dedicated 16-bit hardware. The three test system are as follows:

\begin{itemize}

  \item \emph{Pascal System} - NVIDIA Tesla P100 GPU (16GB), Intel(R) Xeon(R) CPU E5-2690 v4 @ 2.60GHz, 128GB RAM, with Ubuntu 16.04, Python 3.7, CUDA 10.1, CuDNN v7.6 and PyTorch 1.1.

  \item \emph{Turing System} - NVIDIA Titan RTX GPU (24GB), Intel(R) i9-9820X, 64GB RAM, with Arch 5.7.9, Python 3.8, CUDA 10.2, CuDNN v7.6 and PyTorch 1.1.

  \item \emph{Volta System} - NVIDIA Tesla V100 GPU (16GB), Intel(R) Xeon(R) CPU E5-2690 v4 @ 2.60GHz, 128GB RAM, with Ubuntu 16.04, Python 3.7, CUDA 10.1, CuDNN v7.6 and PyTorch 1.1.

\end{itemize}

\subsection{Experiemnts}

Two distinct sets of experiments are performed: firstly, we measure any impact on the ability of the models to make predictions accurately. Secondly, we assess the affect of half-precision on run-time and total GPU memory consumed.

For the predictive performance experiments we measure semi-supervised classification accuracy for the GCN, and Area Under the precision-recall Curve (AUC) and Average Precision (AP) for the link prediction task. These are the performance metrics used to measure the performance of the models when they were introduced \cite{kipf2017semi, kipf2016variational}. The model architecture and primary hyperparameters are fixed and were also taken from the original work and were kept constant across all GPUs, padding use and optimisation levels.

The run time performance results are taken using synthetic graphs so that the size can easily be controlled. This means that all graph sizes are divisible by 8, meaning no padding is required. For these experiments we trained models using the various parameters detailed in Table \ref{tab:hyperparams} and measured both the memory consumption and the total training time. Each combination of parameters from Table \ref{tab:hyperparams} was repeated five times, each repeat with a different random seed.

\begin{table}[h!]
  \centering
  \begin{tabular}{ll}
    \toprule
    \textbf{Parameter} & \textbf{Value Range}\\
    \midrule
    \midrule
		Opt Level         & O0, 01, 02, 03 \\
    Use Features       & False, True \\
    Model Size       & 16, 32, 64, 128, 256, 512, 1024, 2048 \\
    Num Vertices   & $\{x \; | \; 2048 \leq x \leq 2^{15} \equiv 0 \; mod \; 1024\}.$\\
    \bottomrule
	\end{tabular}
  \caption{Model and synthetic data parameter ranges.}
  \label{tab:hyperparams}
  \vskip -15pt
\end{table}
\section{Experimental Results}
\label{sec:results}

In this section, we present the results of our experimental evaluation as discussed in Section \ref{sec:experiment_setup}. We begin by assessing the effects of mixed-precision training on the predictive performance of semi-supervised classification and link prediction. We then move to measure the change in both run-time and the maximum memory consumed on the GPU as the number of vertices in the input graph and the model size is increased for the various levels of reduced-precision optimisation.

\subsection{Assessing Model Predictive Performance}

We first evaluate how predictive performance is affected by the move to reduced precision on the benchmark Cora dataset \cite{yang2016revisiting}. To give a global view of the relationship between the variables, Figure \ref{fig:acc-corr} presents the correlation matrices for both the GCN and GAE approaches on the V100 GPU. It should be noted that very similar results were observed for all cards.

The results demonstrating the performance of the various cards for the GCN model, with and without the use of padding, are presented in Table \ref{tab:gcn-acc-modified-original}. The results in this table measure the classification accuracy on a holdout test set and are presented as the difference, $\Delta$, to the normal full-precision mode, O0. It can be seen that the classification result is within the error margin across all cards for opt levels O2 and O3 -- meaning that mixed-precision training modes can be used without adversely affecting predictive performance. However as expected, the use of O3, complete 16-bit mode, causes a significant drop in accuracy of 64.2\% for the V100 GPU. The results also show that the padding applied to the graph has no significant impact on model accuracy.

\begin{table}[t!]
  \centering

  \begin{tabular}{l c c c}
  \toprule
  \multicolumn{1}{l}{\multirow{2}{*}{\textbf{GPU}}} & \multicolumn{1}{c}{\multirow{2}{*}{\textbf{Opt Level}}} & \multicolumn{2}{c}{$\Delta$ \textbf{Accuracy}}\T\B\\

  \cline{3-4}
          &	& \textbf{w/ Padding}	& \textbf{w/o Padding}\T\B\\

  \midrule \midrule

\multirow{3}{*}{V100} & O1   & + 0.005 & - 0.005\\
                      & O2   & + 0.005 & - 0.003\\
                      & O3   & - 0.642 & - 0.648\\

\midrule

\multirow{3}{*}{Titan RTX} & O1   & + 0.001 & + 0.002\\
                           & O2   & + 0.003 & + 0.004\\
                           & O3   & - 0.647 & - 0.646\\

\midrule

\multirow{4}{*}{P100}     & O1   & + 0.002  & - 0.001\\
                          & O2   & + 0.004  & 0\\
                          & O3   & - 0.642  & 0.637\\
\bottomrule
\end{tabular}

\caption{Comparison of the GCN classification results on the Cora dataset using vertex features across GPUs and optimisation levels. All elements indicate the difference $\Delta$ between the values from Ox and O0.}
\label{tab:gcn-acc-modified-original}
\end{table}


The results for the task of link prediction using the GAE model are presented in Table \ref{tab:gae-acc}. The results largely conform to those presented for the GCN, with the use of O1 and O2 having no significant impact on predictive performance versus the full precision baseline, and O3 causing significant degradation. Again, it can be seen that the use of padding does not impact performance.

\begin{table}[h!]
  \centering
  \resizebox{\linewidth}{!}{

  \begin{tabular}{@{\extracolsep{4pt}}l c c c c c}
  \toprule
  \multicolumn{1}{l}{\multirow{2}{*}{\textbf{GPU}}} & \multicolumn{1}{c}{\multirow{2}{*}{\textbf{Opt Level}}} & \multicolumn{2}{c}{\textbf{w/ Padding}} & \multicolumn{2}{c}{\textbf{w/o Padding}}\T\B\\

  \cline{3-4} \cline{5-6}
          &	& $\Delta$ \textbf{AUC}	& $\Delta$ \textbf{AP} & $\Delta$ \textbf{AUC}	& $\Delta$ \textbf{AP}\T\B\\

  \midrule \midrule

\multirow{3}{*}{V100}       & O1   & - 0.005 & - 0.006 & + 0.001 & - 0.005\\
                            & O2   & 0      & 0      & + 0.001 & 0\\
                            & O3   & - 0.277 & - 0.195 & - 0.276 & - 0.194\\

\midrule

\multirow{3}{*}{Titan RTX}  & O1   & - 0.005  & - 0.007 & - 0.005 & - 0.006\\
                            & O2   & 0       &  0     & 0      & 0\\
                            & O3   & - 0.277  & - 0.195 & - 0.276  & - 0.194\\

\midrule

\multirow{3}{*}{P100}       & O1   & - 0.005  & - 0.007 & - 0.005 & - 0.006\\
                            & O2   & 0       & 0      & 0      & 0\\
                            & O3   & - 0.277  & - 0.195 & - 0.276 & - 0.194\\

\bottomrule
\end{tabular}
}
\caption{Comparison of GAE edge prediction results on the Cora dataset using vertex features across GPUs and optimisation levels. All elements indicate the difference $\Delta$ between the values from Ox and O0.}
\label{tab:gae-acc}
\end{table}

\subsection{Run-Time and Memory Usage Analysis}

In this section, we present the results measuring how run-time and memory usage change as both the size of the input Barab\'{a}si-Albert graph and the model size are altered. For all results in this section, unless otherwise stated, the identity matrix of the graph is used as the input features. Additionally, the figures are presented as the mean over five model seeds and either all model sizes, for the case of the graph size plots, or all graph sizes, for the case of the model size plots. Error bars are presented as the standard deviation of these.

\begin{figure*}[t]
  \centering
    \begin{subfigure}[b]{0.28\textwidth}
      \includegraphics[width=\textwidth]{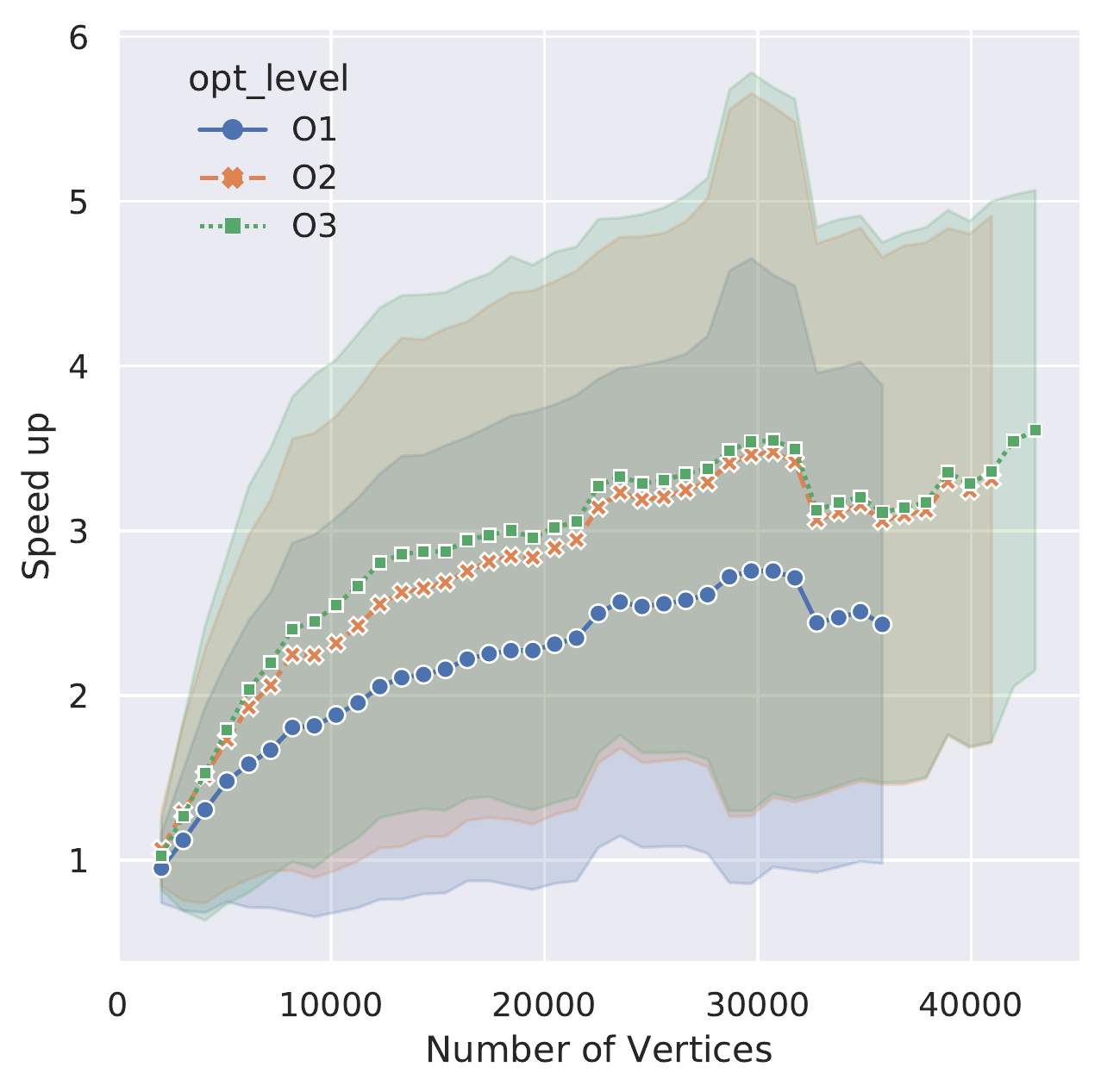}
      \caption{Titan RTX}
    \end{subfigure}
    \begin{subfigure}[b]{0.29\textwidth}
      \includegraphics[width=\textwidth]{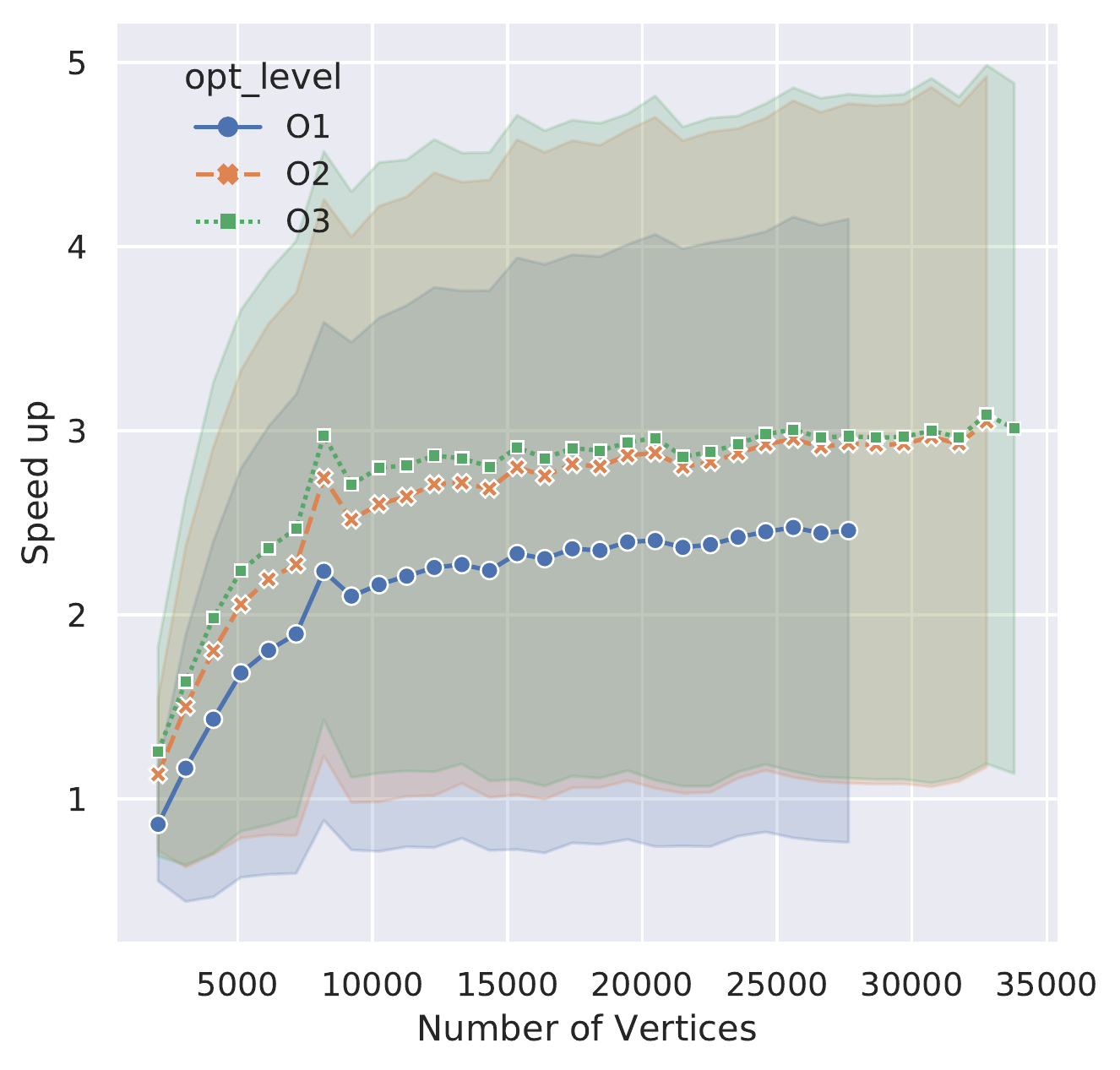}
      \caption{V100}
    \end{subfigure}
    \begin{subfigure}[b]{0.28\textwidth}
      \includegraphics[width=\textwidth]{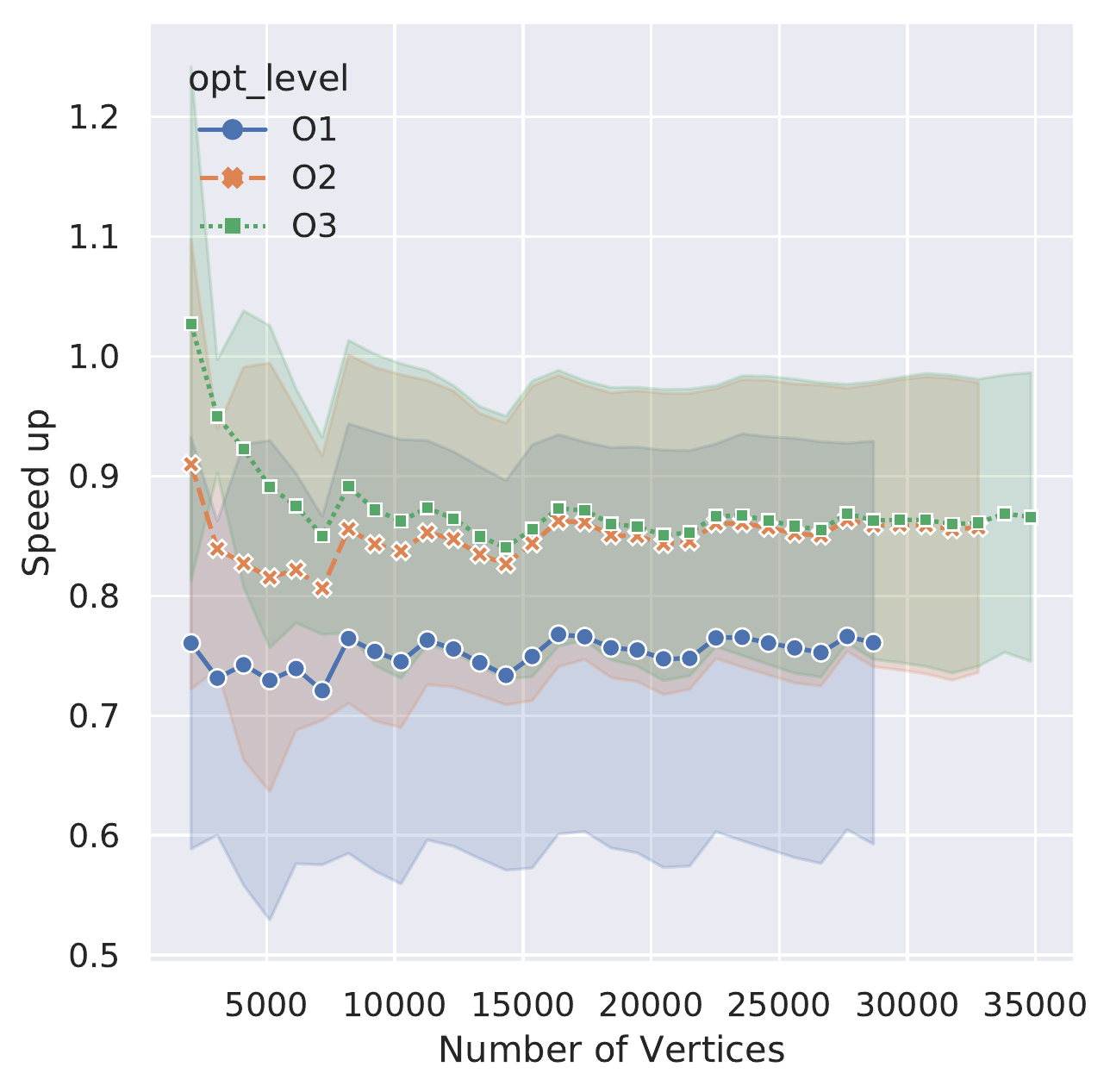}
      \caption{P100}
      \label{fig:gcn-vertex-speedup-p100}
    \end{subfigure}
    \caption{Speed up of the various opt levels versus O0 for the GCN approach.}
    \label{fig:gcn-vertex-speedup}
\end{figure*}

\begin{figure*}[t]
  \centering
    \begin{subfigure}[b]{0.28\textwidth}
      \includegraphics[width=\textwidth]{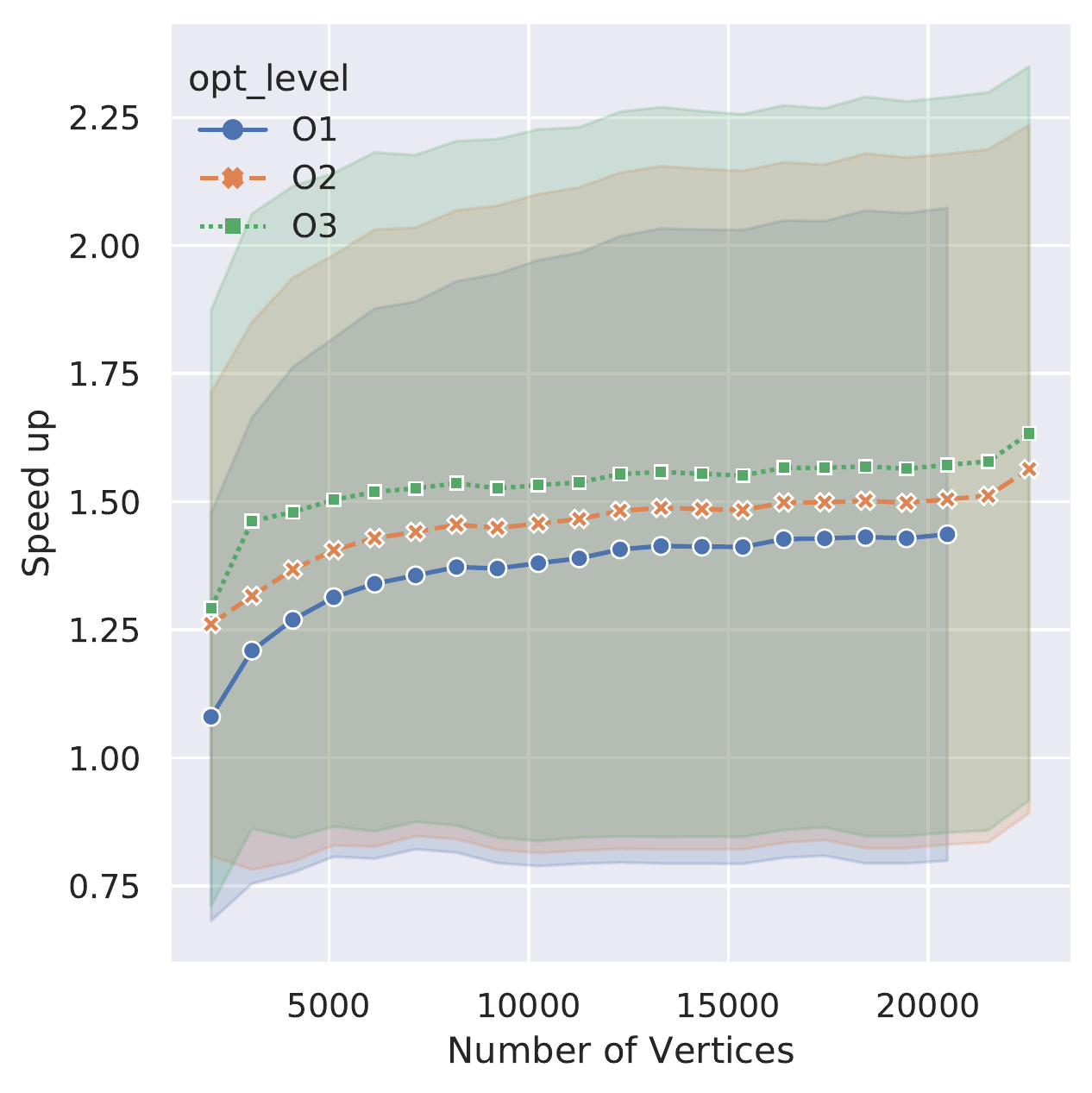}
      \caption{Titan RTX}
    \end{subfigure}
    \begin{subfigure}[b]{0.29\textwidth}
      \includegraphics[width=\textwidth]{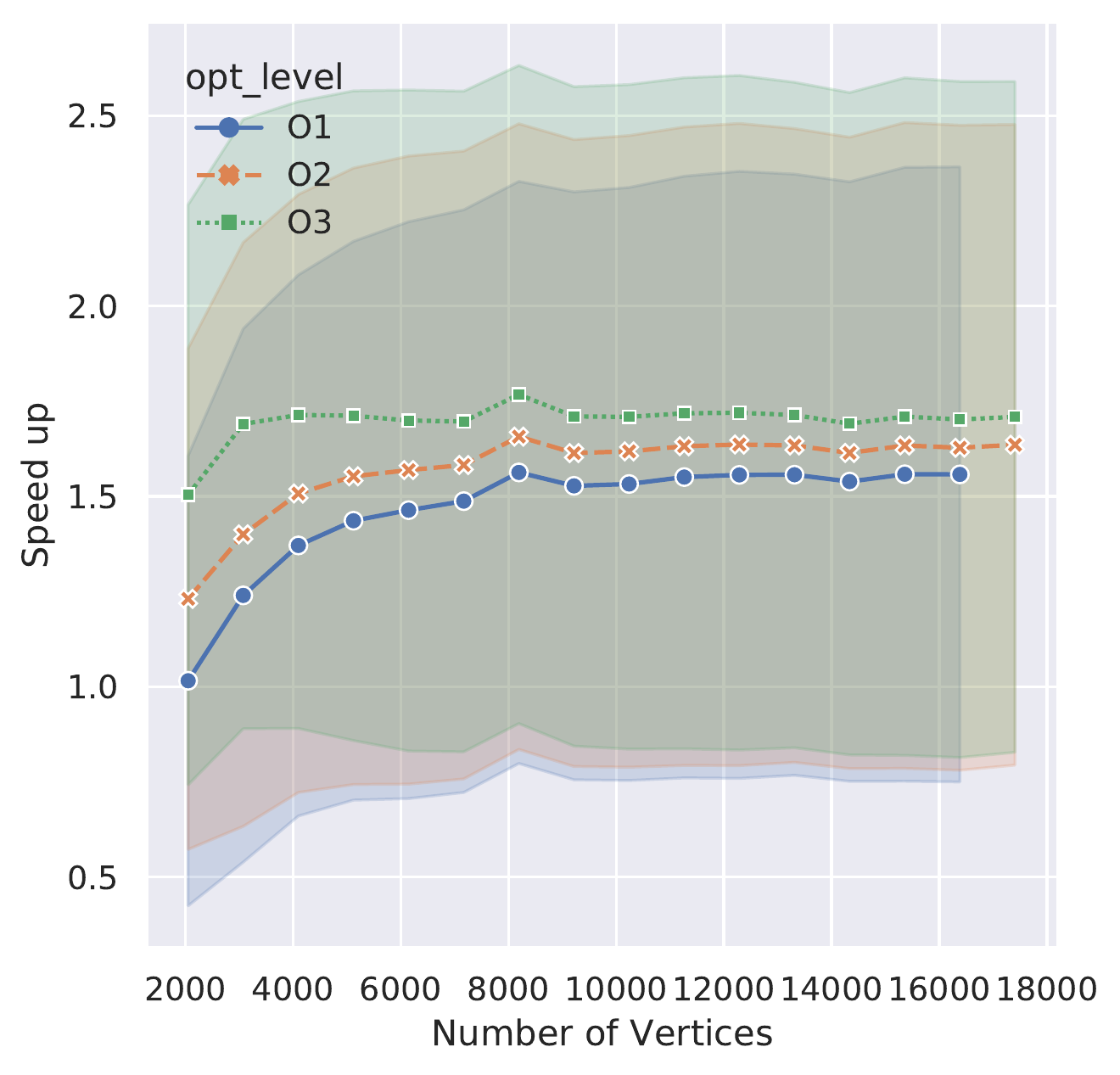}
      \caption{V100}
    \end{subfigure}
    \begin{subfigure}[b]{0.28\textwidth}
      \includegraphics[width=\textwidth]{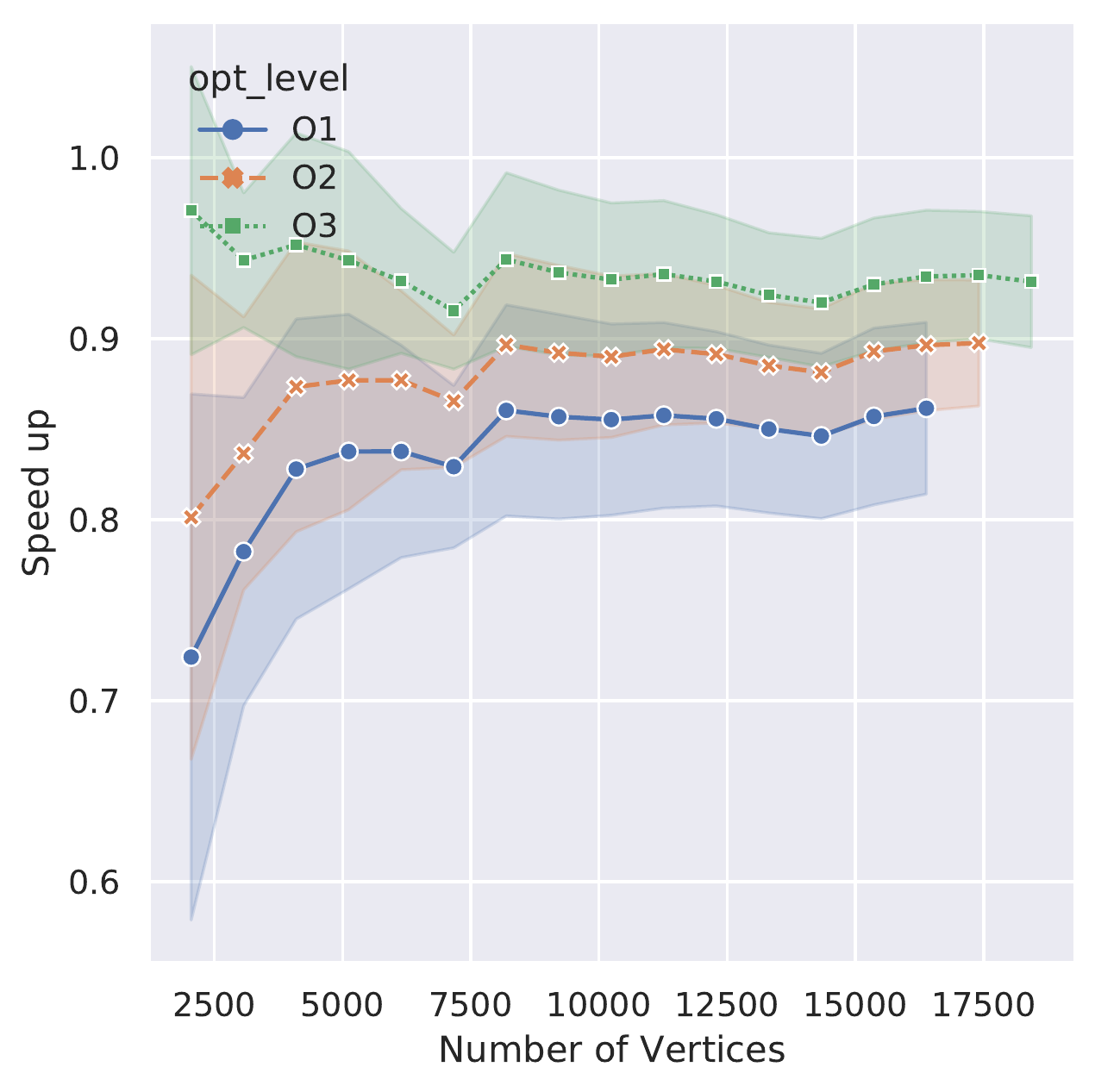}
      \caption{P100}
    \end{subfigure}
    \caption{Speed up of the various opt levels versus O0 for the GAE approach.}
    \label{fig:gae-vertex-speedup}
\end{figure*}

To give an overview of how the various factors are related, Figure \ref{fig:mem-corr} presents a correlation matrix for the run-time experiments for both the GCN and GAE models running on the V100 GPU. Some unsurprising observations can be made across both model types, for example the positive correlation between the model size and the total training time. However, some perhaps unexpected ones also arise - there is a clear positive correlation between the training run being killed because of an Out Of Memory (OOM) error and the use of opt level O1.

\subsubsection{Measuring Run-Time and Memory Usage Versus Graph Size}

We now focus on measuring how the run-time and memory usage are affected as the number of vertices in the input training graph is changed. Figure \ref{fig:gcn-vertex-runtime} demonstrates how the run-time of the GCN model responds as the number of vertices in the input graph is increased. Firstly, by comparing between the two cards with Tensor Cores (Figures \ref{fig:gcn-vertex-runtime-rtx} and \ref{fig:gcn-vertex-runtime-v100}) and the one without dedicated 16-bit hardware (Figure \ref{fig:gcn-vertex-runtime-p100}), the effect at opt levels O1-O3 is immediately obvious. The P100 results in Figure \ref{fig:gcn-vertex-runtime-p100} show the opt level to have almost no impact on the training time, whereas the other cards show a clear decrease in run-time whenever a 16-bit mode is enabled. One interesting observation is that the standard deviation (error bar) of each point is much lower for the 16-bit opt levels. As each point is presented as the mean over seeds and all the model sizes for that graph size, this would indicate that opt levels O1-O3 are less sensitive to model size when compared to the graph size. This will be investigated further in the next section when changes with respect to model size are studied. Perhaps the most surprising result is that opt level O0, the full 32-bit training model, can scale to larger graph sizes than the other levels. This is of note as intuitively one would expect that the reduced-precision modes would use less memory, and thus to scale to a larger input dataset size -- however this is clearly not the case. Further evidence of this will be presented when we consider memory usage.

To further investigate how various opt levels affect performance relative to the baseline of full-precision, we present the speed up of all opt levels relative to O0 for the GCN approach in Figure \ref{fig:gcn-vertex-speedup}. The figure highlights how mixed-precision can offer large speed ups for graph-based neural models, with the proviso that the GPU has dedicated hardware support for the operations. Figure \ref{fig:gcn-vertex-speedup-p100} demonstrates that using mixed precision when the GPU is lacking the 16-bit specific hardware can actually result in a worse run-time overall, as illustrated by the speed up value of below 1 across all opt levels.

\begin{figure*}[t]
  \centering
    \begin{subfigure}[b]{0.28\textwidth}
      \includegraphics[width=\textwidth]{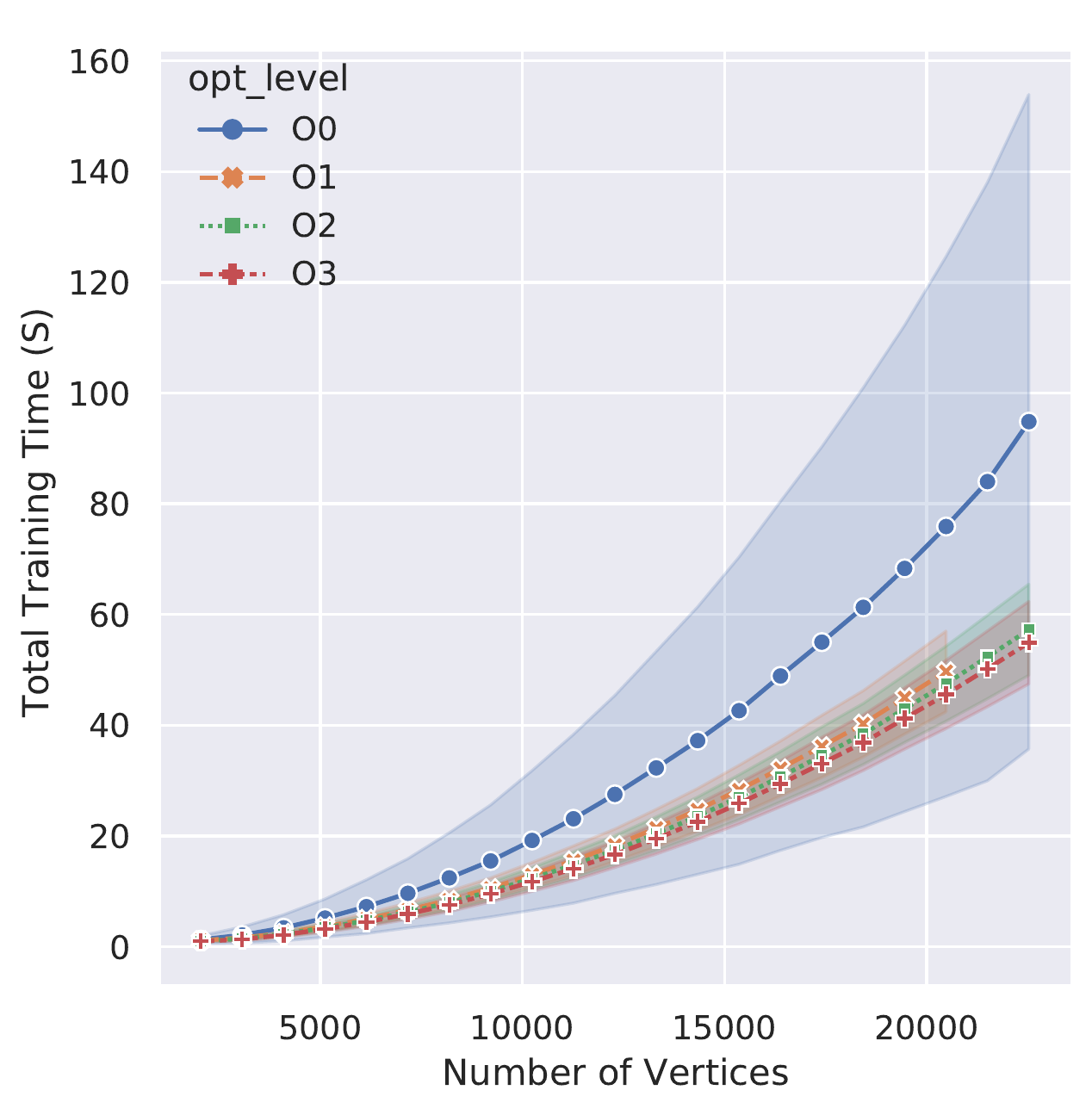}
      \caption{Titan RTX}
    \end{subfigure}
    \begin{subfigure}[b]{0.292\textwidth}
      \includegraphics[width=\textwidth]{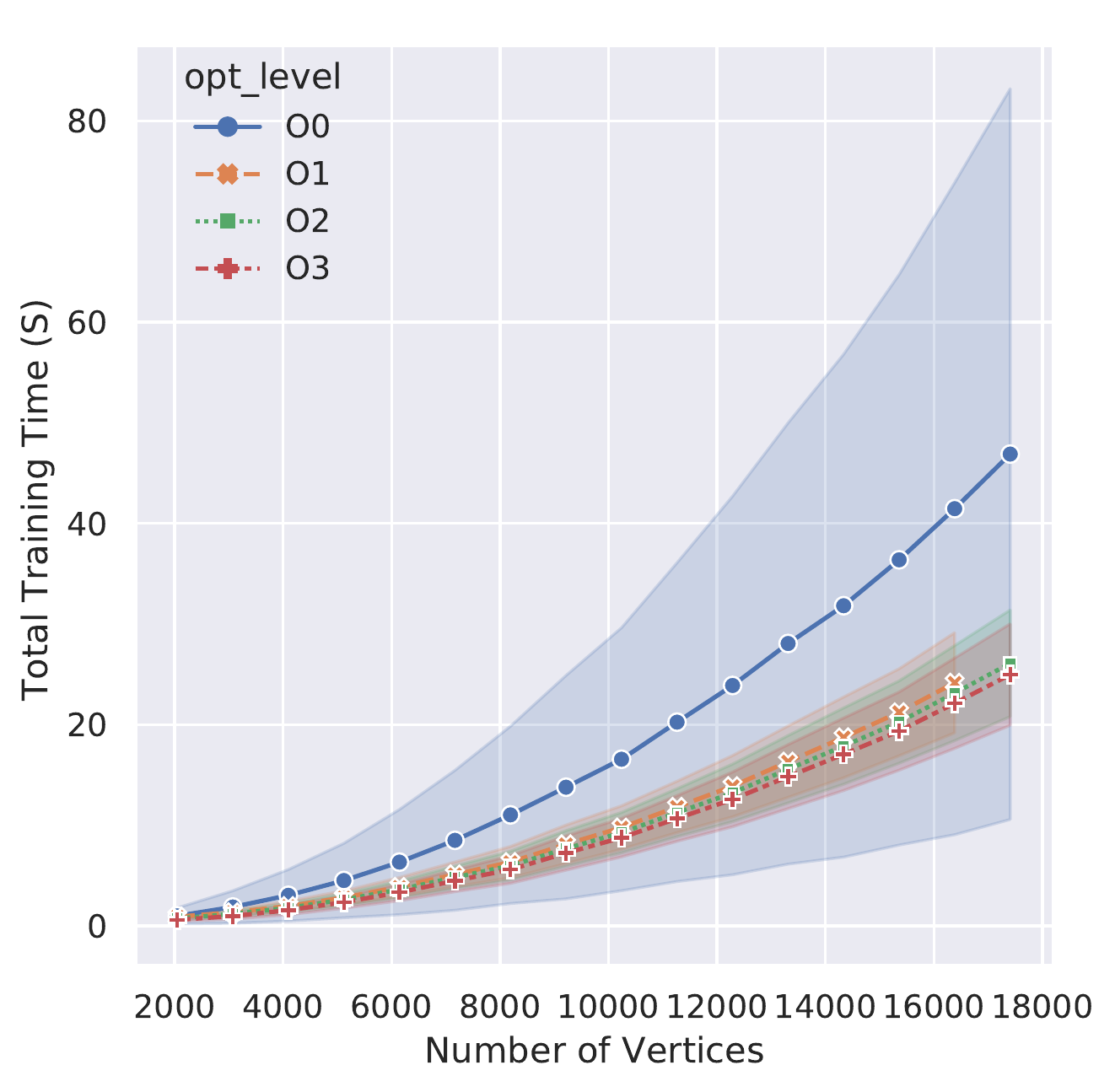}
      \caption{V100}
    \end{subfigure}
    \begin{subfigure}[b]{0.28\textwidth}
      \includegraphics[width=\textwidth]{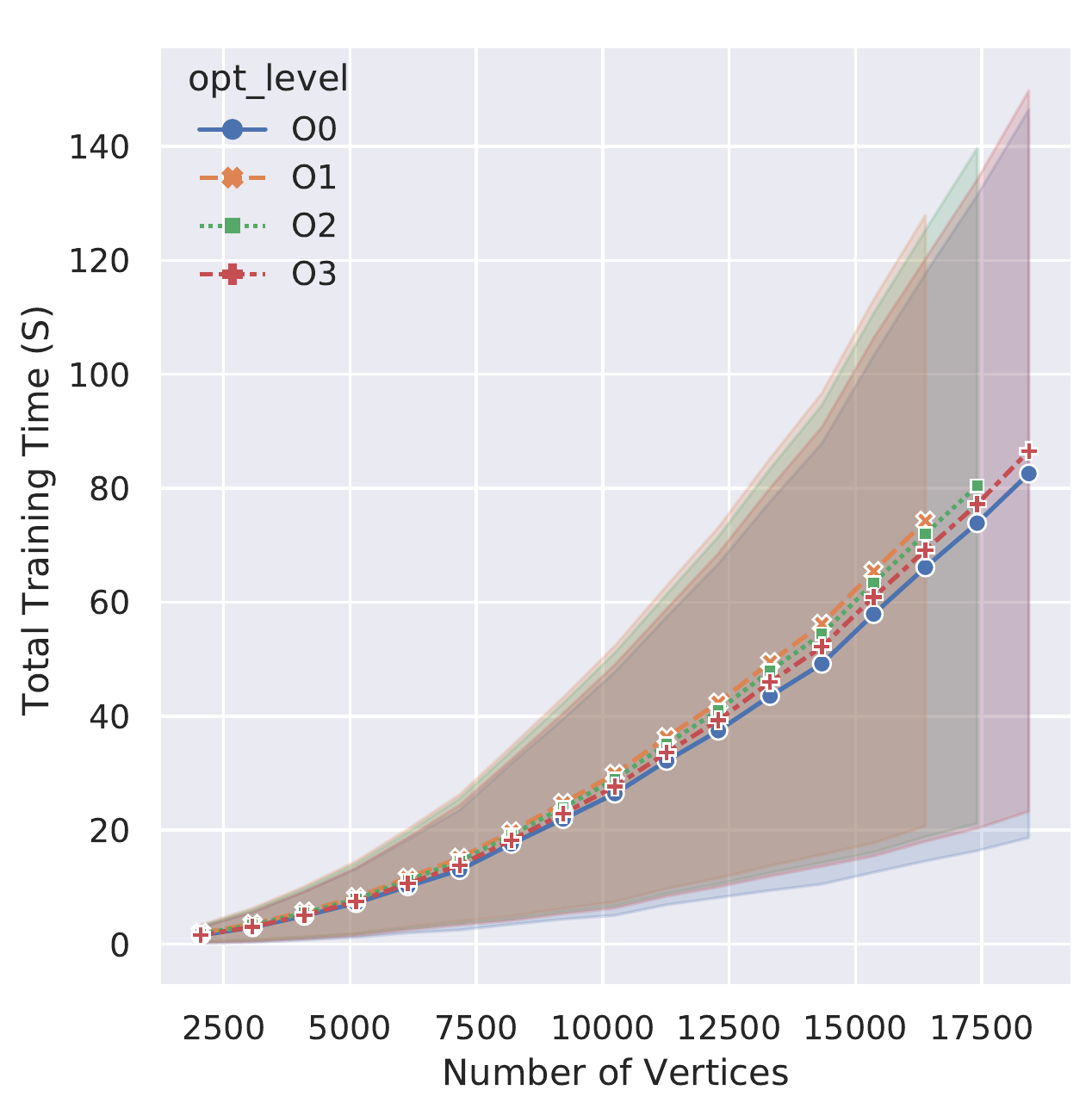}
      \caption{P100}
      \label{fig:gae-vertex-runtime-p100}
    \end{subfigure}
    \caption{GAE total training time versus increases in graph size.}
    \label{fig:gae-vertex-runtime}
\end{figure*}

\begin{figure}
  \centering
  \begin{subfigure}[b]{0.24\textwidth}
  \includegraphics[width=\textwidth]{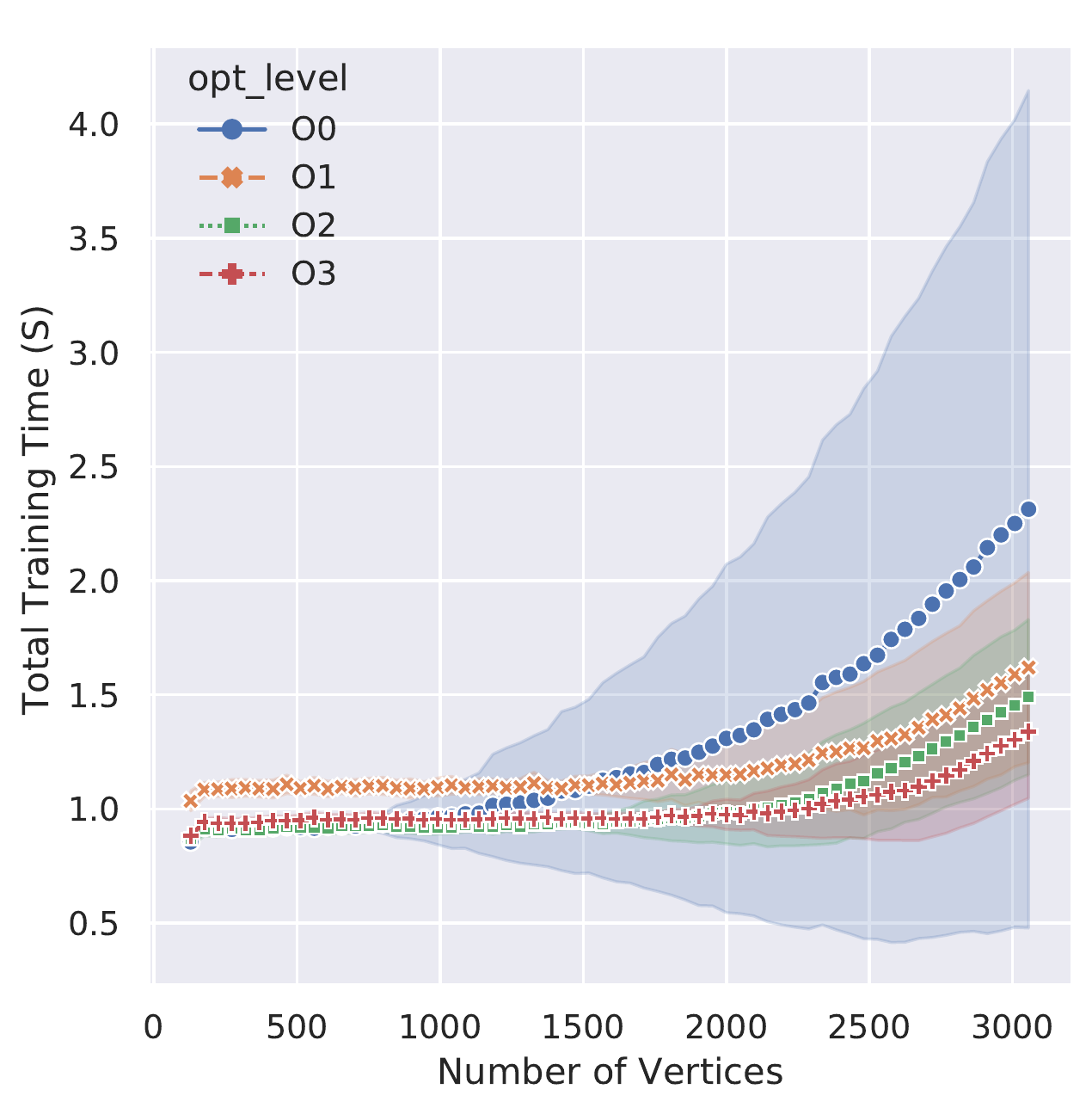}
  \caption{Total Time}
  \label{fig:gae_zoomed_time}
  \end{subfigure}
  \begin{subfigure}[b]{0.24\textwidth}
    \includegraphics[width=\textwidth]{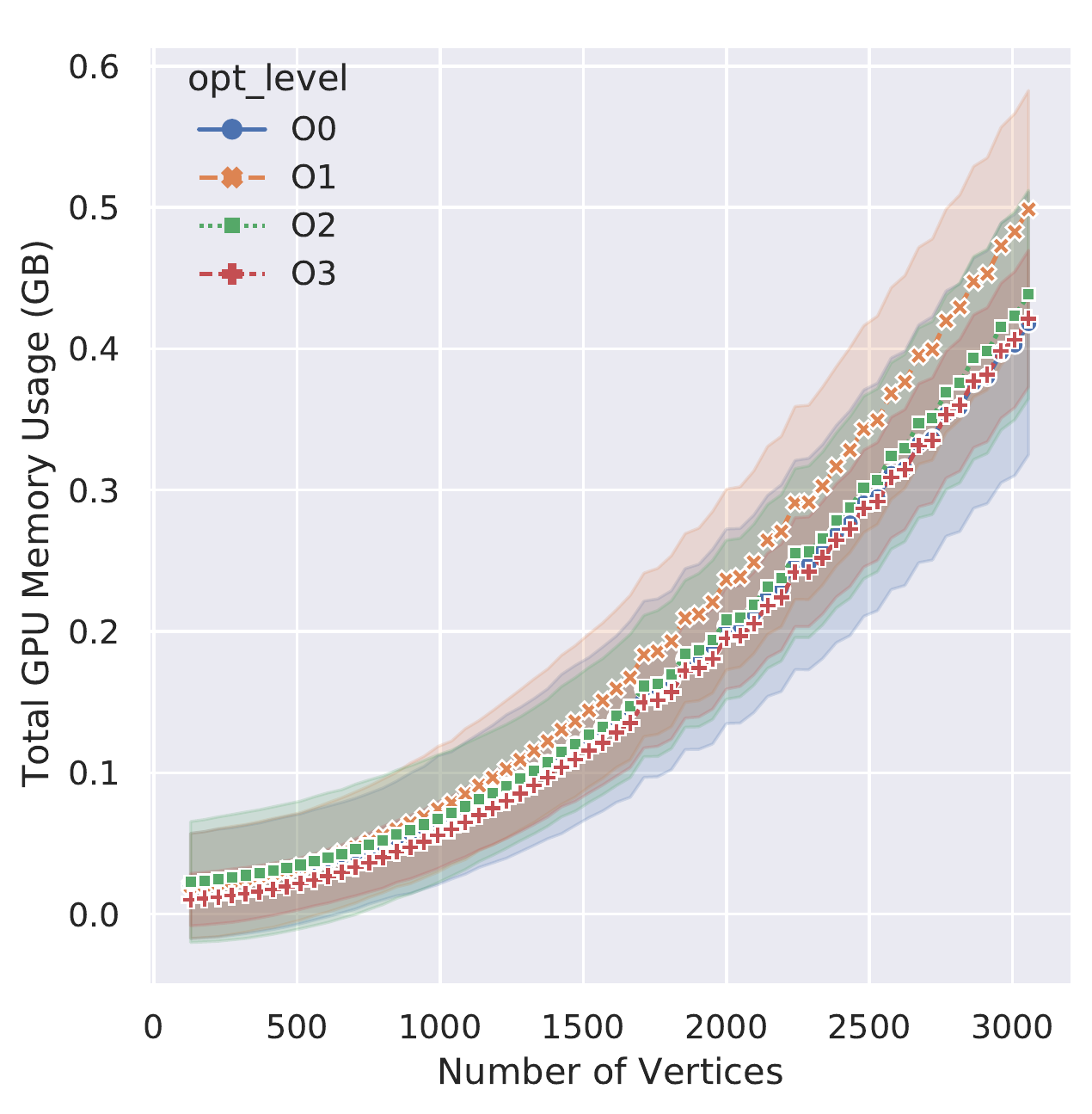}
  \caption{Mem Usage}
  \label{fig:gae_zoomed_mem}
  \end{subfigure}
  \caption{A truncated view of the GAE results for the V100.}
  \label{fig:gae_zoomed}
\end{figure}

We also highlight how the GAE model is affected by the various optimisation levels, with Figure \ref{fig:gae-vertex-runtime} showing the change in total training time versus the input graph size. We can again see in Figure \ref{fig:gae-vertex-runtime-p100} that using a mixed-precision training mode offers no run-time benefit if the GPU lacks dedicated hardware support. However, one clear trend is that, when compared to the GCN model, the GAE does not show the same level of decrease in run-time when mixed-precision training is utilised. This is demonstrated by opt level O0 being much closer to the mixed-precision modes, although it is still significantly higher. Another continuing trend however is the much lower run-time variance over model sizes for the mixed-precision approaches. To investigate at what graph size the mixed-precision modes start to outperform the baseline, we present a truncated view of the GAE results for the V100 in Figure \ref{fig:gae_zoomed_time}. The figure shows that at a graph size of 1,000 vertices, opt levels O2 and O3 start to outperform O0, with O1 also outperforming it when 1,500 vertices is reached.

Figure \ref{fig:gae-vertex-speedup} highlights the speed up from using mixed-precision modes with the GAE model. It shows that the speed up is on average less than what was shown with the GCN approach. We suspect this is due to the much more complex graph reconstruction loss function required by the GAE approach, as this may not benefit much from the use of reduced precision. Another interesting trend in the figure is that the speed up is much more consistent across graph sizes, with the average speed up being almost identical from 4,000 to 18,000 vertices.

\begin{figure*}[h]
  \centering
    \begin{subfigure}[b]{0.28\textwidth}
      \includegraphics[width=\textwidth]{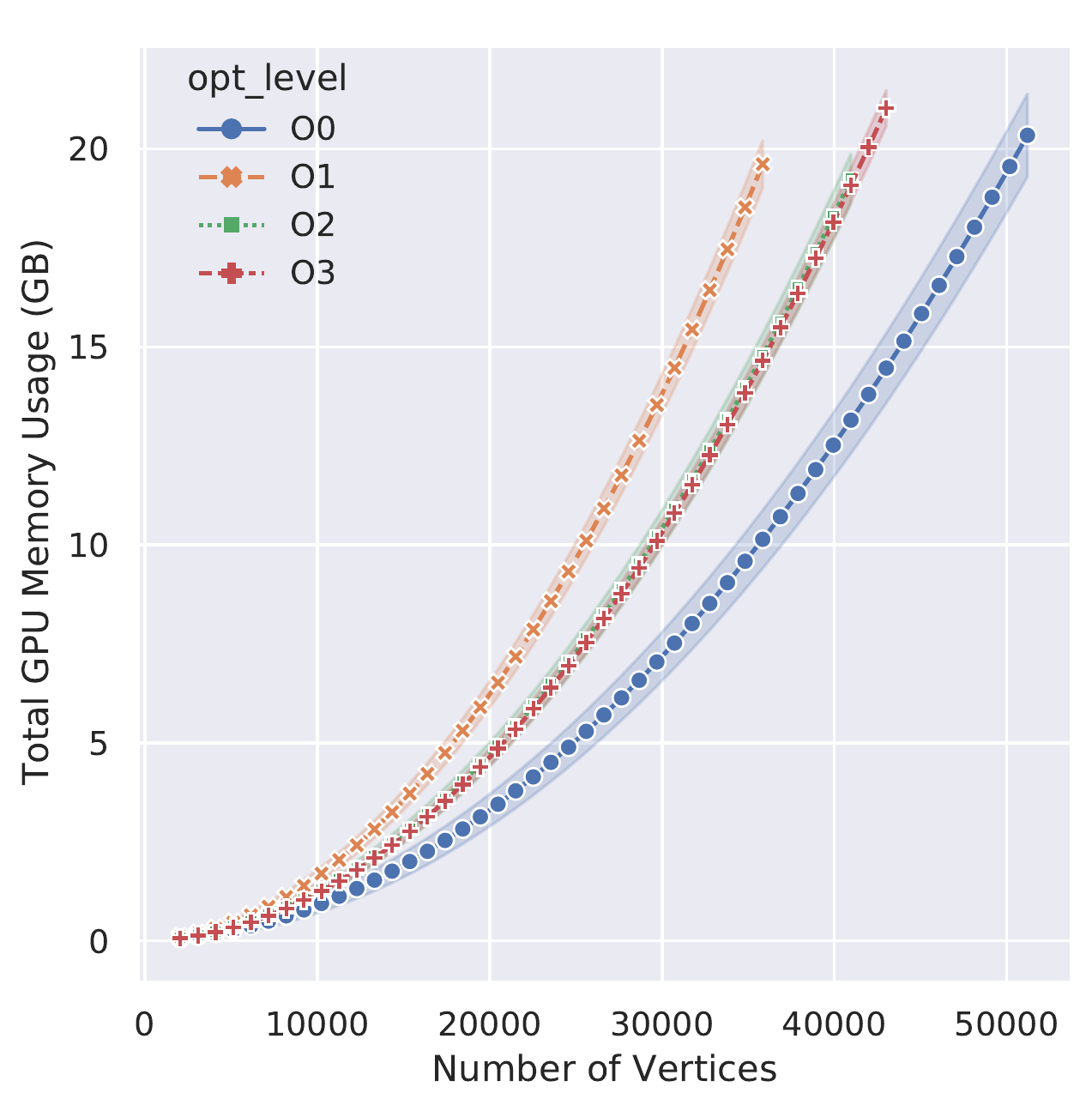}
      \caption{Titan RTX}
    \end{subfigure}
    \begin{subfigure}[b]{0.285\textwidth}
      \includegraphics[width=\textwidth]{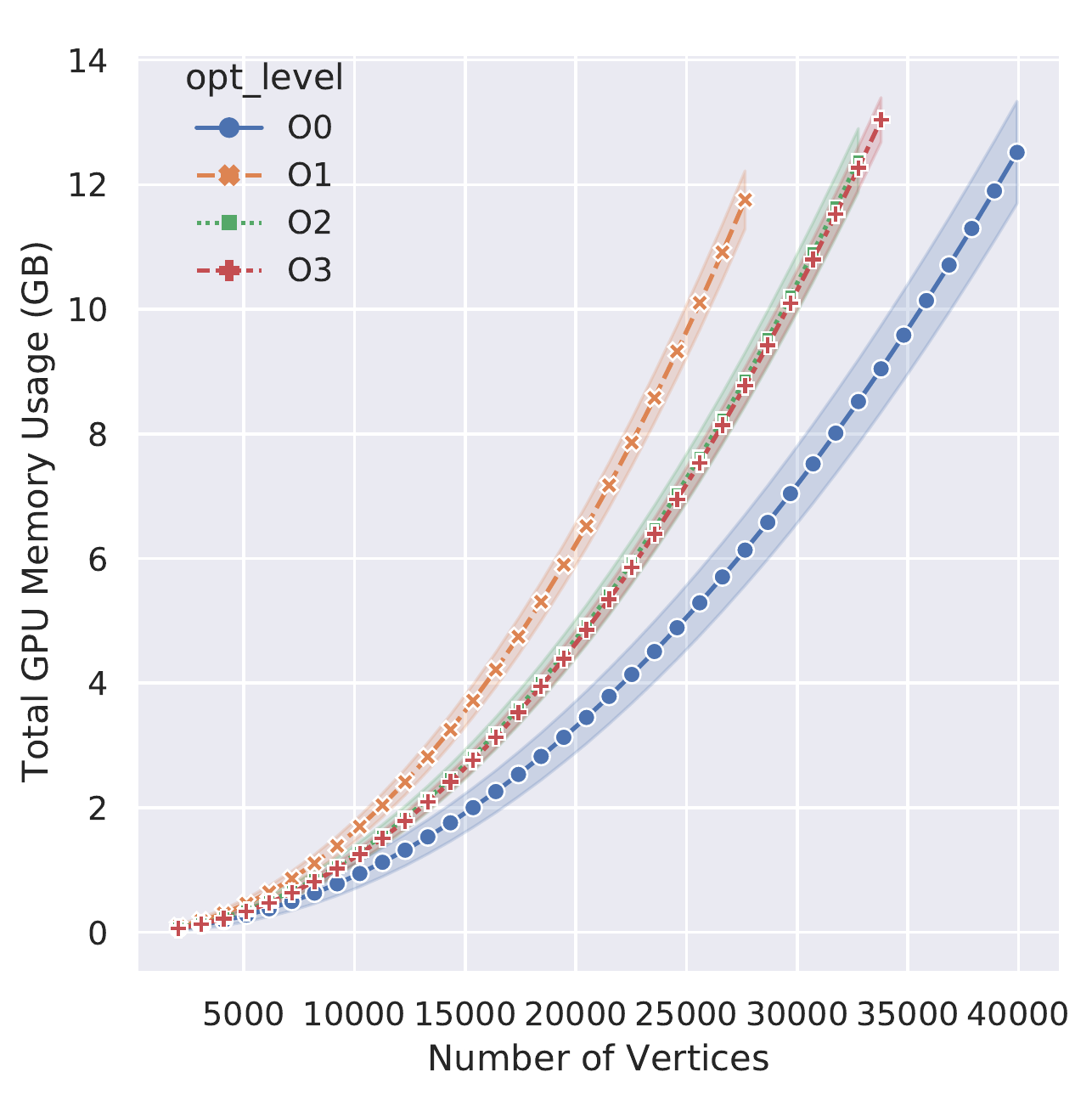}
      \caption{V100}
    \end{subfigure}
    \begin{subfigure}[b]{0.28\textwidth}
      \includegraphics[width=\textwidth]{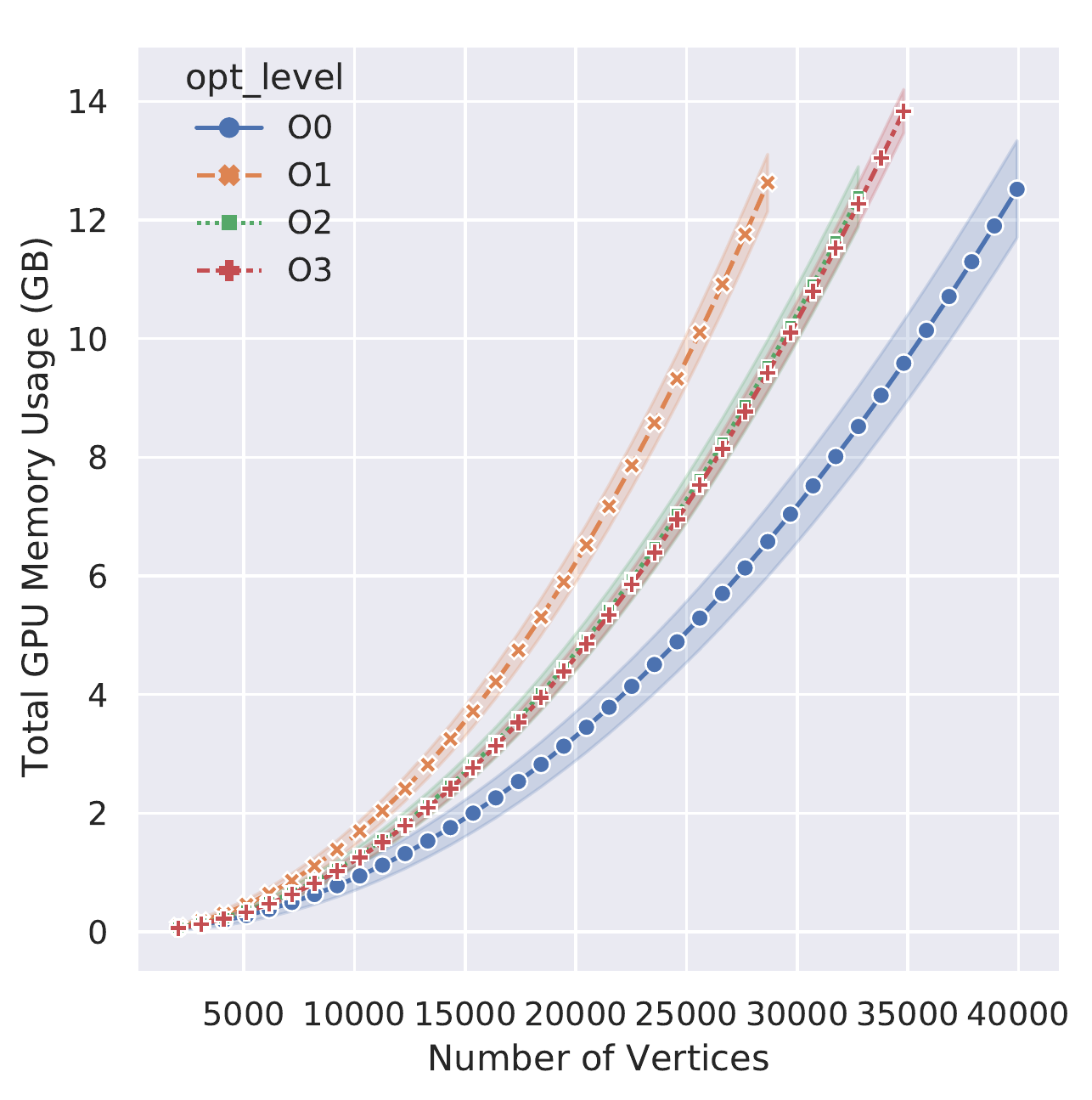}
      \caption{P100}
    \end{subfigure}
    \caption{Maximum amount of GPU memory consumed during the training process across all cards for the GCN model.}
    \label{fig:gcn-vertex-mem}
\end{figure*}

Our experiments also include an analysis of how the change in the input graph size affects the maximum memory consumed on the GPU during the training process, with results across the three GPUs being presented in Figure \ref{fig:gcn-vertex-mem}. There is one clear and perhaps unintuitive result demonstrated across all cards - \emph{using a reduced-precision mode of any kind results in more memory usage when compared to full precision for a given graph size.} The result explains the earlier observation that using a reduced-precision mode means that a smaller total graph size can run when compared to opt level O0 -- they were running out of available memory sooner. It is interesting to note that the results are highly consistent across all cards, demonstrating that even when offering no performance benefit on the P100, the reduced-precision modes O1-O2 are still consuming the additional memory.

\begin{figure*}[h]
  \centering
    \begin{subfigure}[b]{0.28\textwidth}
      \includegraphics[width=\textwidth]{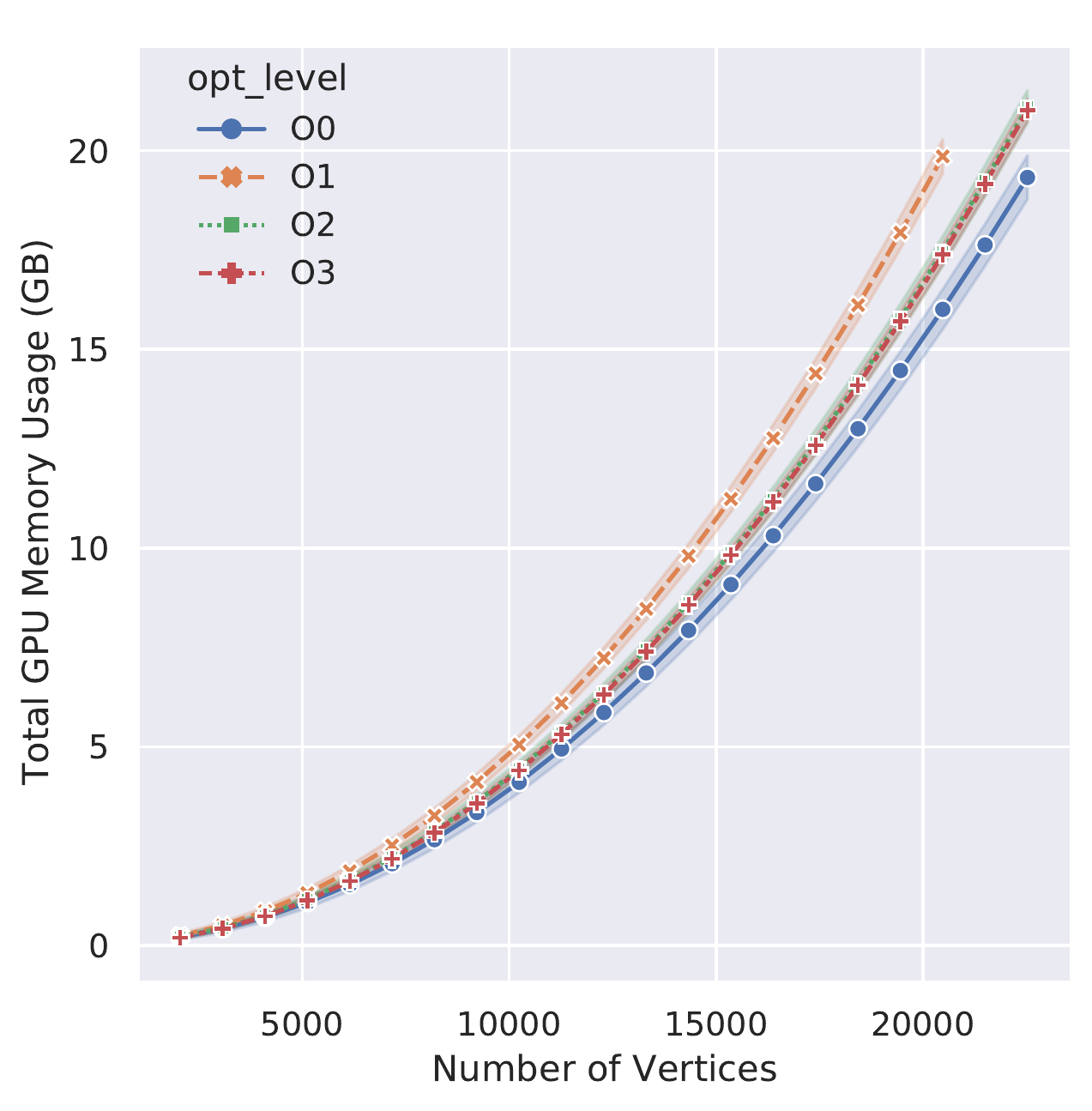}
      \caption{Titan RTX}
    \end{subfigure}
    \begin{subfigure}[b]{0.29\textwidth}
      \includegraphics[width=\textwidth]{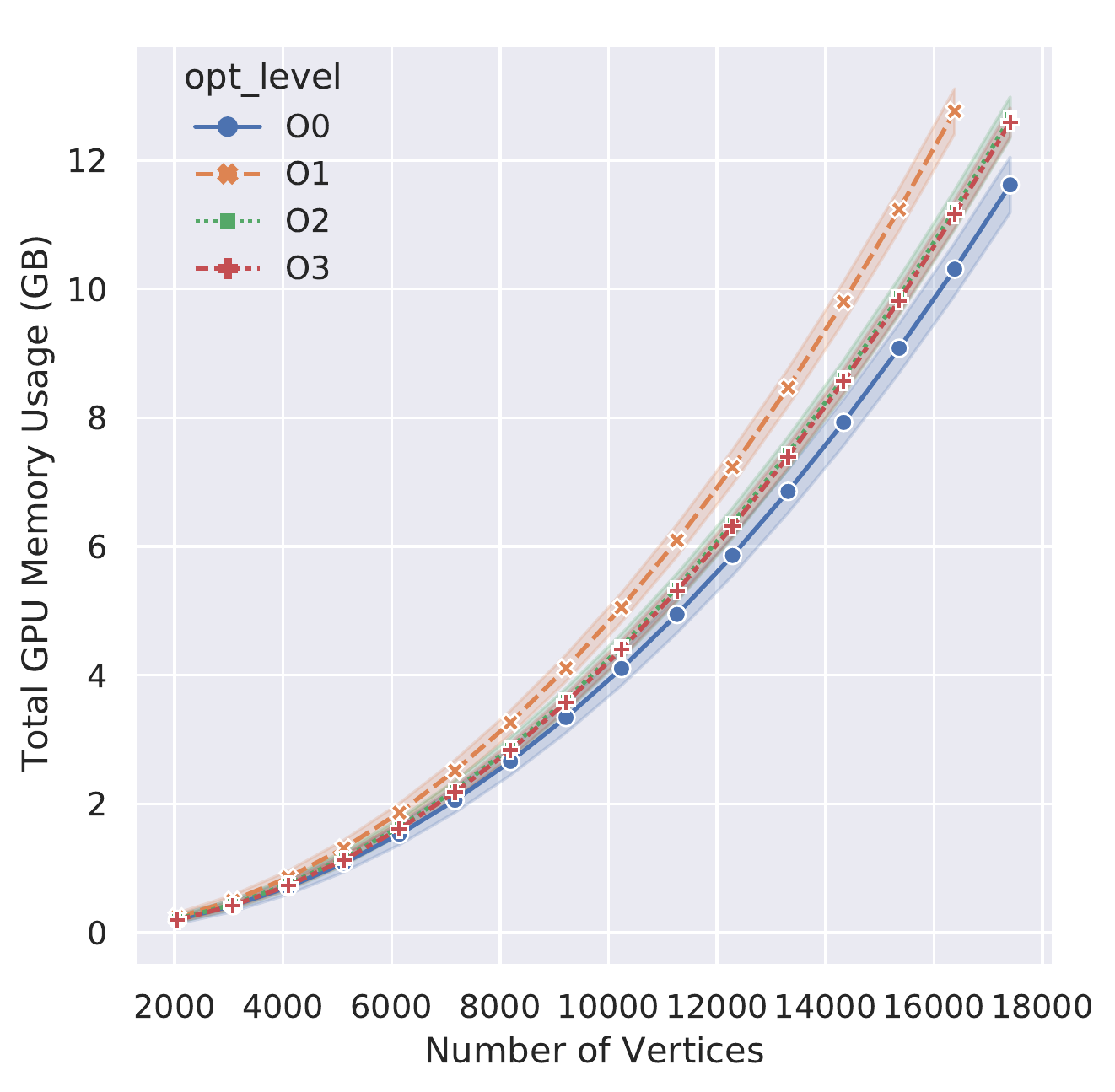}
      \caption{V100}
    \end{subfigure}
    \begin{subfigure}[b]{0.28\textwidth}
      \includegraphics[width=\textwidth]{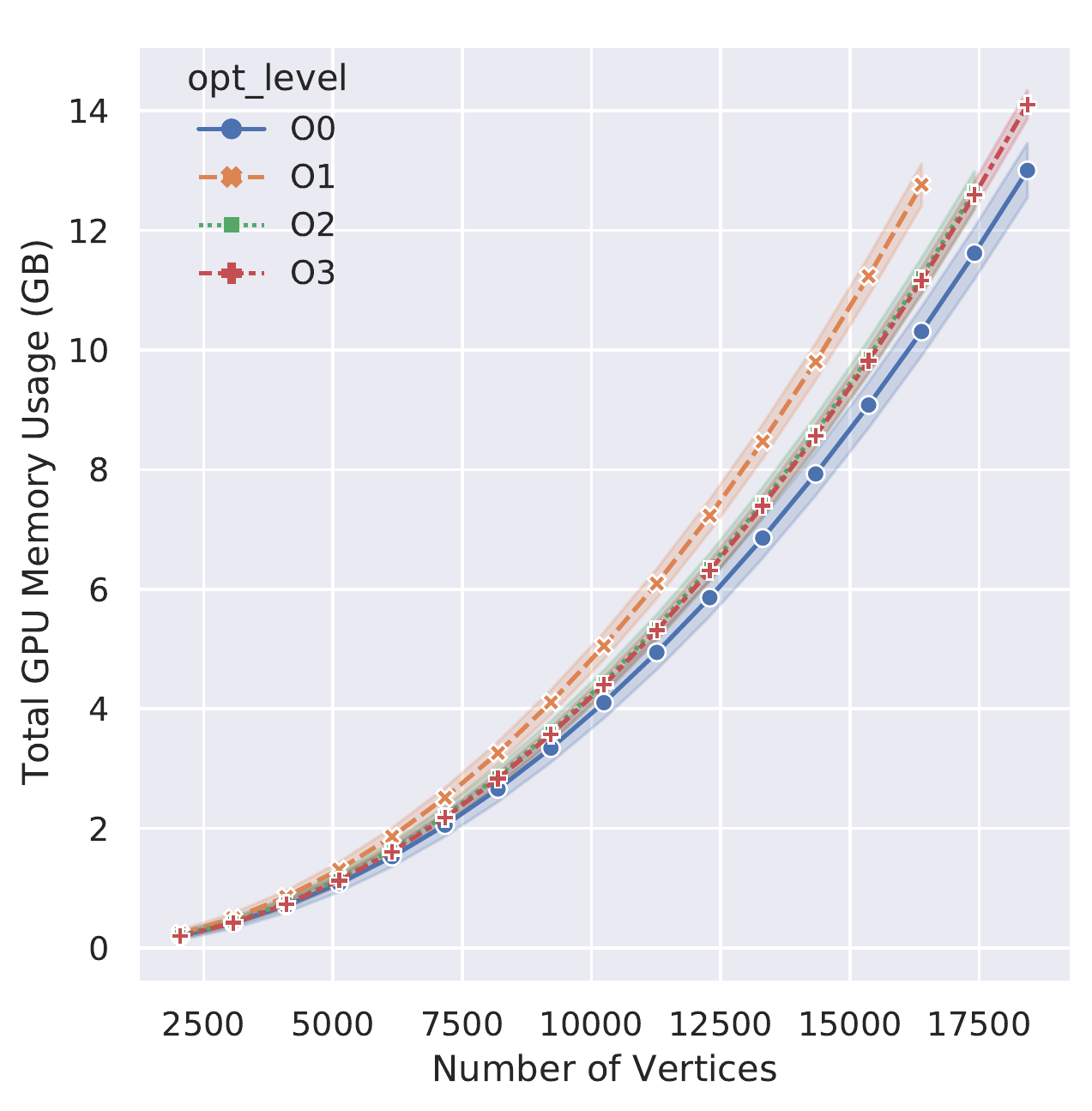}
      \caption{P100}
    \end{subfigure}
    \caption{GAE max memory usage versus increases in graph size.}
    \label{fig:gae-model-mem}
\end{figure*}

Figure \ref{fig:gae-model-mem} highlights how the maximum memory required for the various opt levels changes with respect to the number of vertices in the input graph. The figure highlights how, as was true for the run-time, the mixed-precision opt levels are much closer together with respect to the baseline here. It also re-emphasises how similar the pattern of memory usage is across the various cards -- even on the P100 GPU. To analyse at what point the memory usage of the mixed-precision approaches starts to increase, Figure \ref{fig:gae_zoomed_mem} presents a truncated view of the memory usage for the V100 GPU. The figure shows by a graph size of 1,000 vertices, opt level O1 is starting to demonstrate more memory usage than the other optimisation levels.

\subsubsection{Measuring Run-Time and Memory Usage Versus Model Size}

\begin{figure*}[h]
  \centering
    \begin{subfigure}[b]{0.28\textwidth}
      \includegraphics[width=\textwidth]{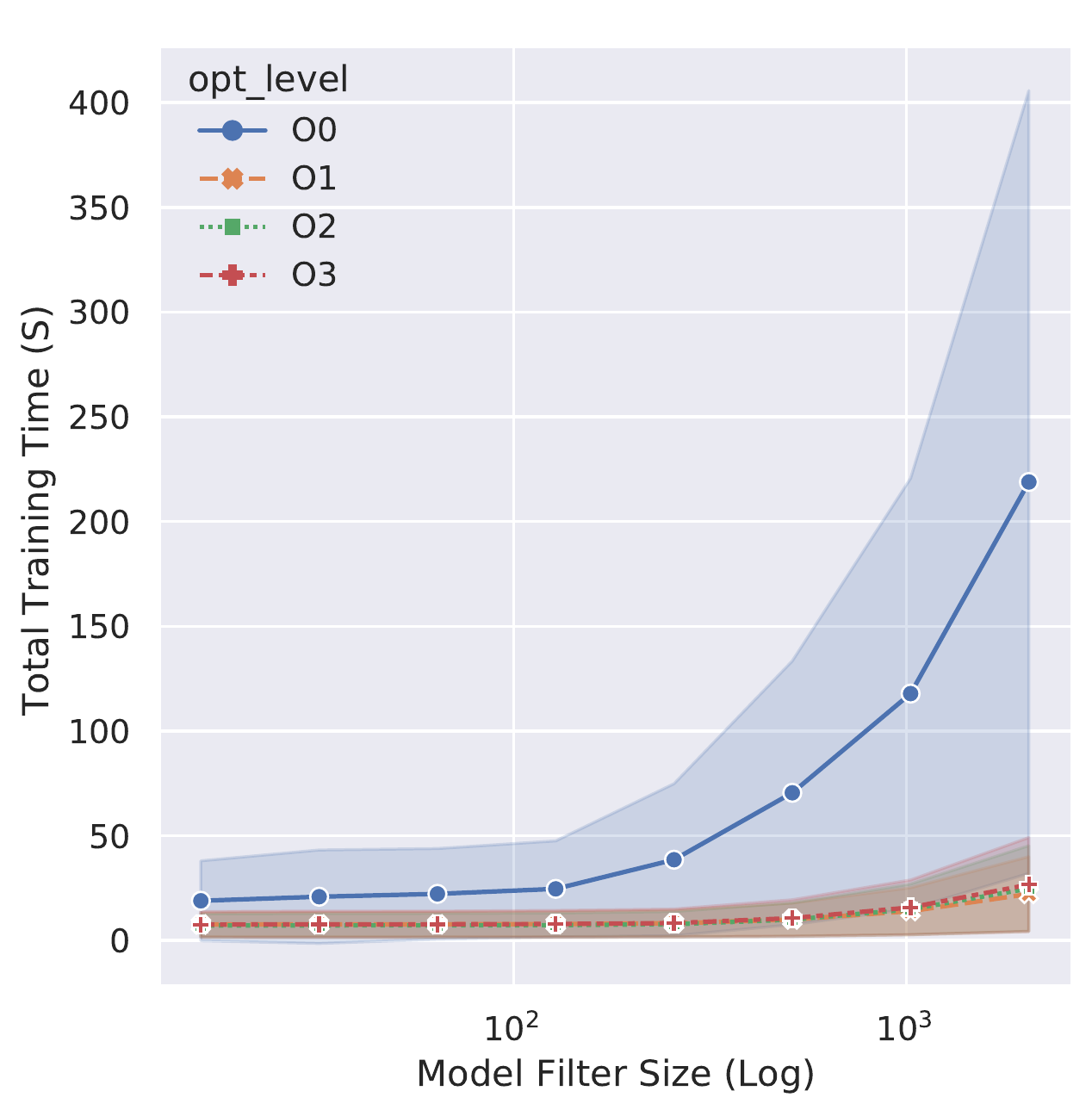}
      \caption{Titan RTX}
    \end{subfigure}
    \begin{subfigure}[b]{0.28\textwidth}
      \includegraphics[width=\textwidth]{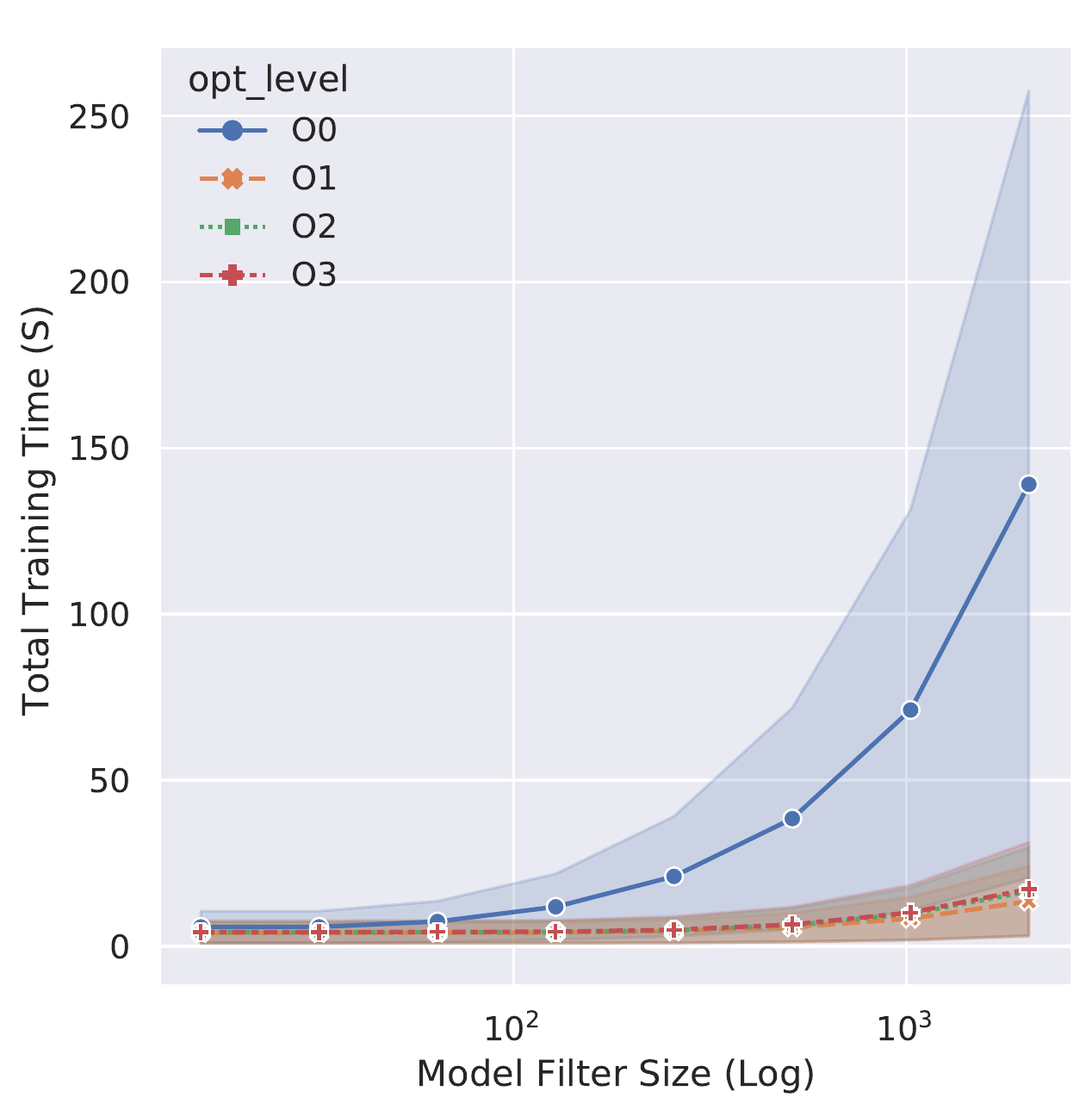}
      \caption{V100}
    \end{subfigure}
    \begin{subfigure}[b]{0.28\textwidth}
      \includegraphics[width=\textwidth]{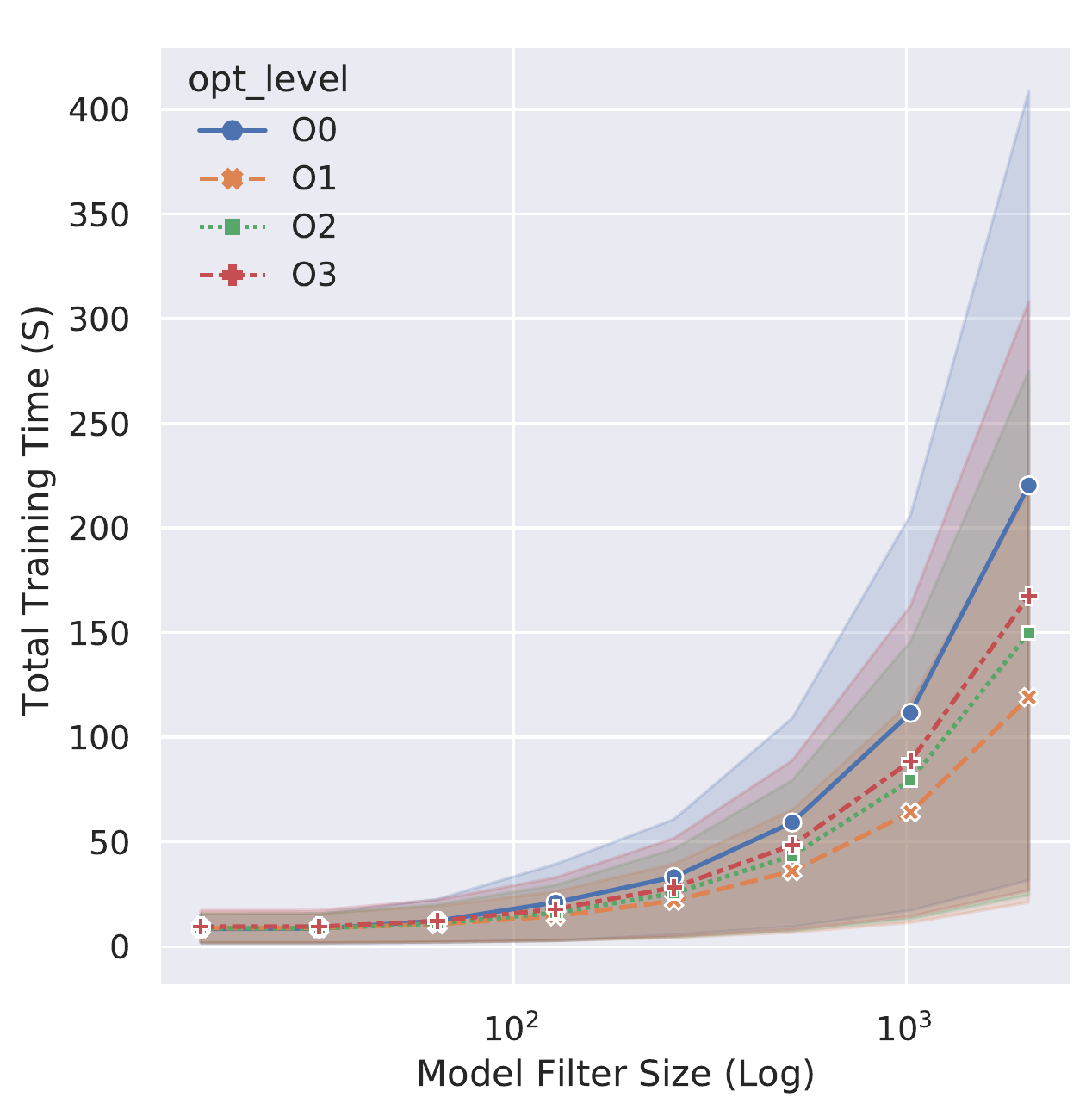}
      \caption{P100}
    \end{subfigure}
    \caption{Total training time as the model size is increased for the GCN model.}
    \label{fig:gcn-model-time}
\end{figure*}

\begin{figure*}[h]
  \centering
    \begin{subfigure}[b]{0.28\textwidth}
      \includegraphics[width=\textwidth]{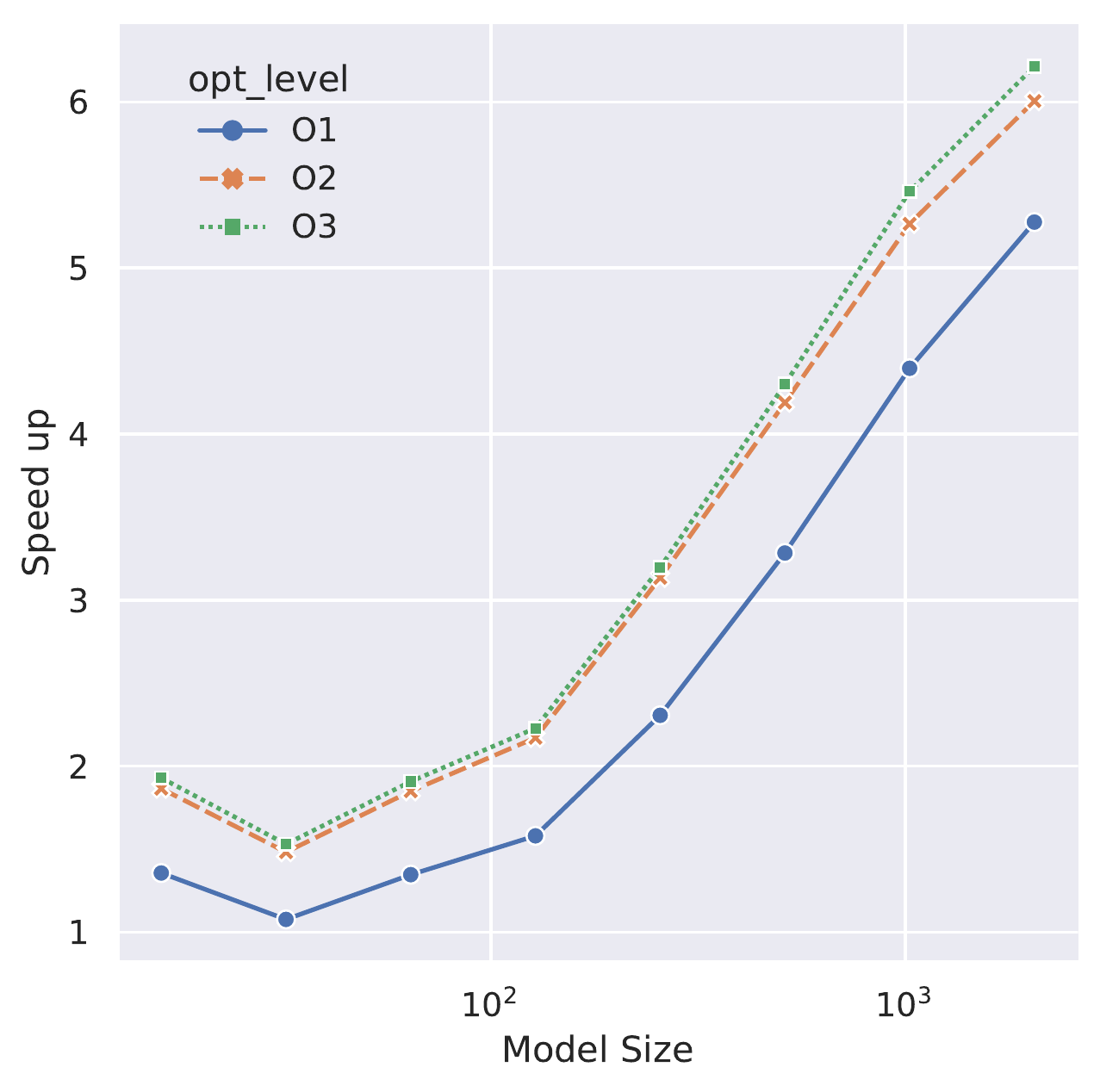}
      \caption{Titan RTX}
    \end{subfigure}
    \begin{subfigure}[b]{0.28\textwidth}
      \includegraphics[width=\textwidth]{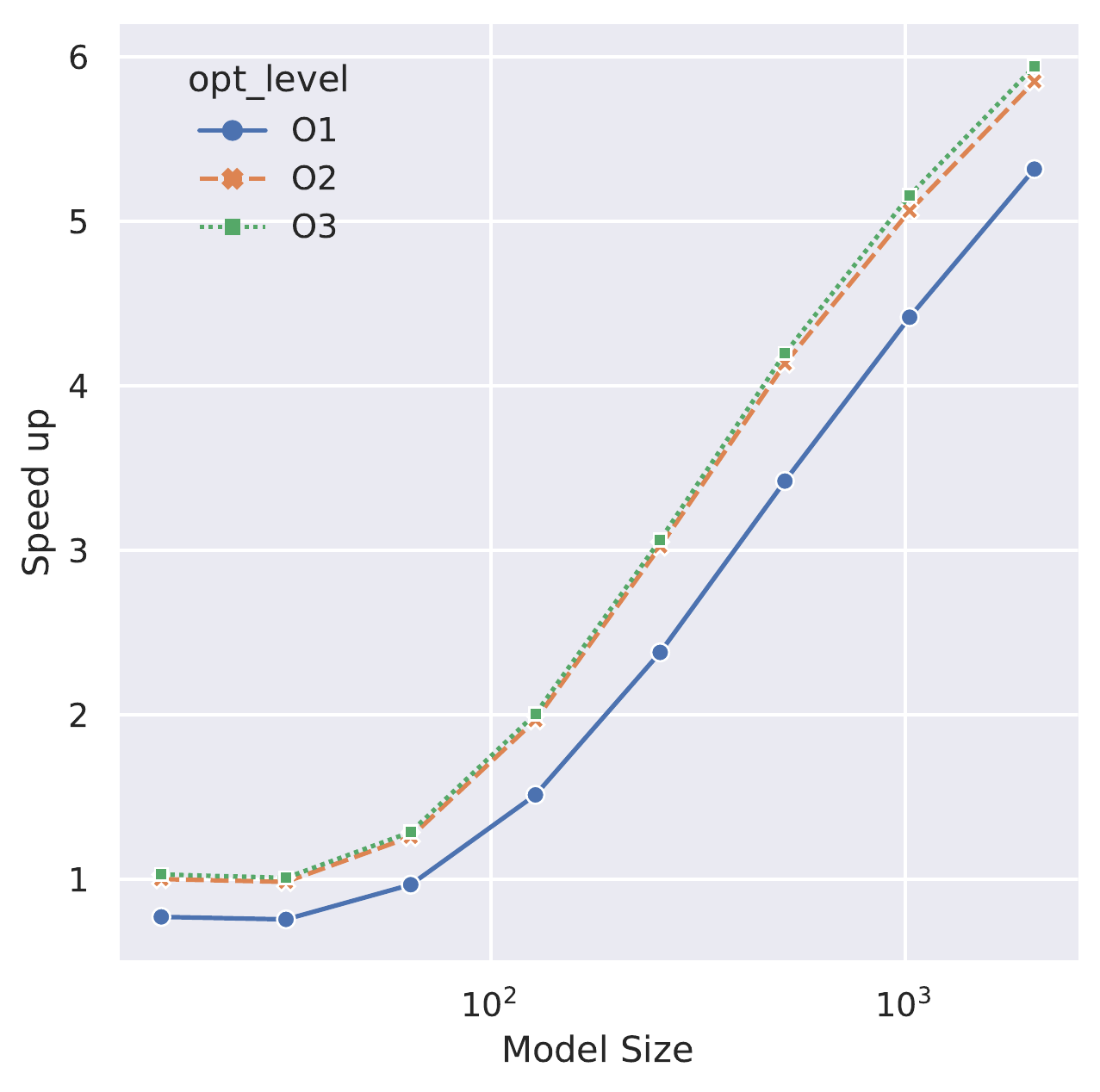}
      \caption{V100}
    \end{subfigure}
    \begin{subfigure}[b]{0.28\textwidth}
      \includegraphics[width=\textwidth]{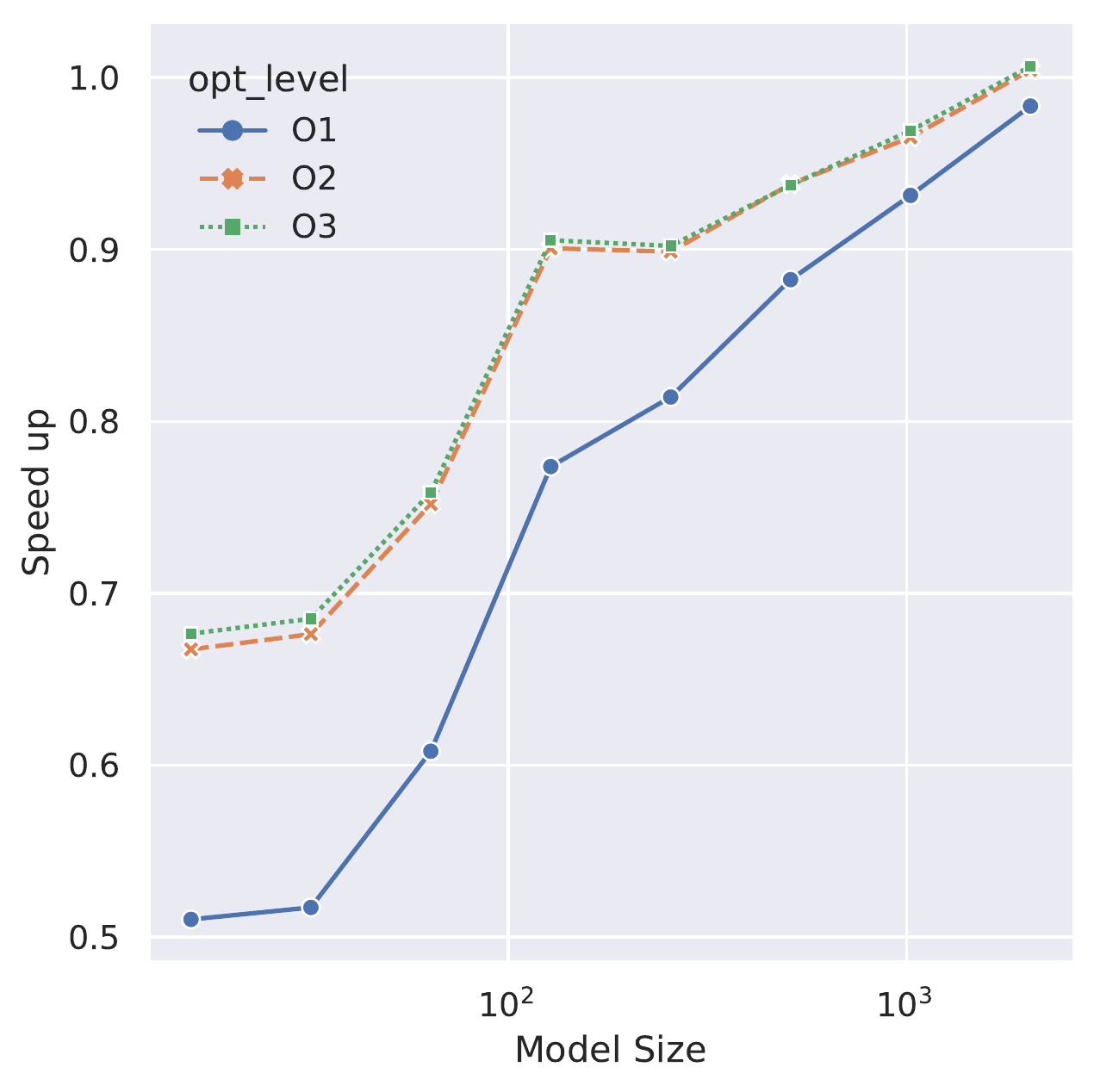}
      \caption{P100}
    \end{subfigure}
    \caption{Speed up of the various opt levels versus O0 for the GCN approach. Results presented using the large graph size that was able to complete with all model sizes}
    \label{fig:gcn-model-speedup}
\end{figure*}

We now present results demonstrating how the performance and memory consumption is affected by the model size. Figure \ref{fig:gcn-model-time} shows the increases in model sizes against the total training time. Conforming to the trend established earlier, not using reduced precision results in large increases in run-time as larger model sizes are used. Conversely, any use of reduced precision means that the run-time is largely unaffected by increases in the number of model parameters -- a very interesting observation. Figure \ref{fig:gcn-model-speedup} demonstrates the speed up of the various reduced-precision optimisation levels against the full-precision baseline. The figure shows that the potential speed up by using reduced precision continues to increase as large model sizes are used, indeed the speed up has not plateaued even with the largest size used. This suggests that further experiments could be run to determine at what point the speed up no longer increases.

\begin{figure*}[h]
  \centering
    \begin{subfigure}[b]{0.28\textwidth}
      \includegraphics[width=\textwidth]{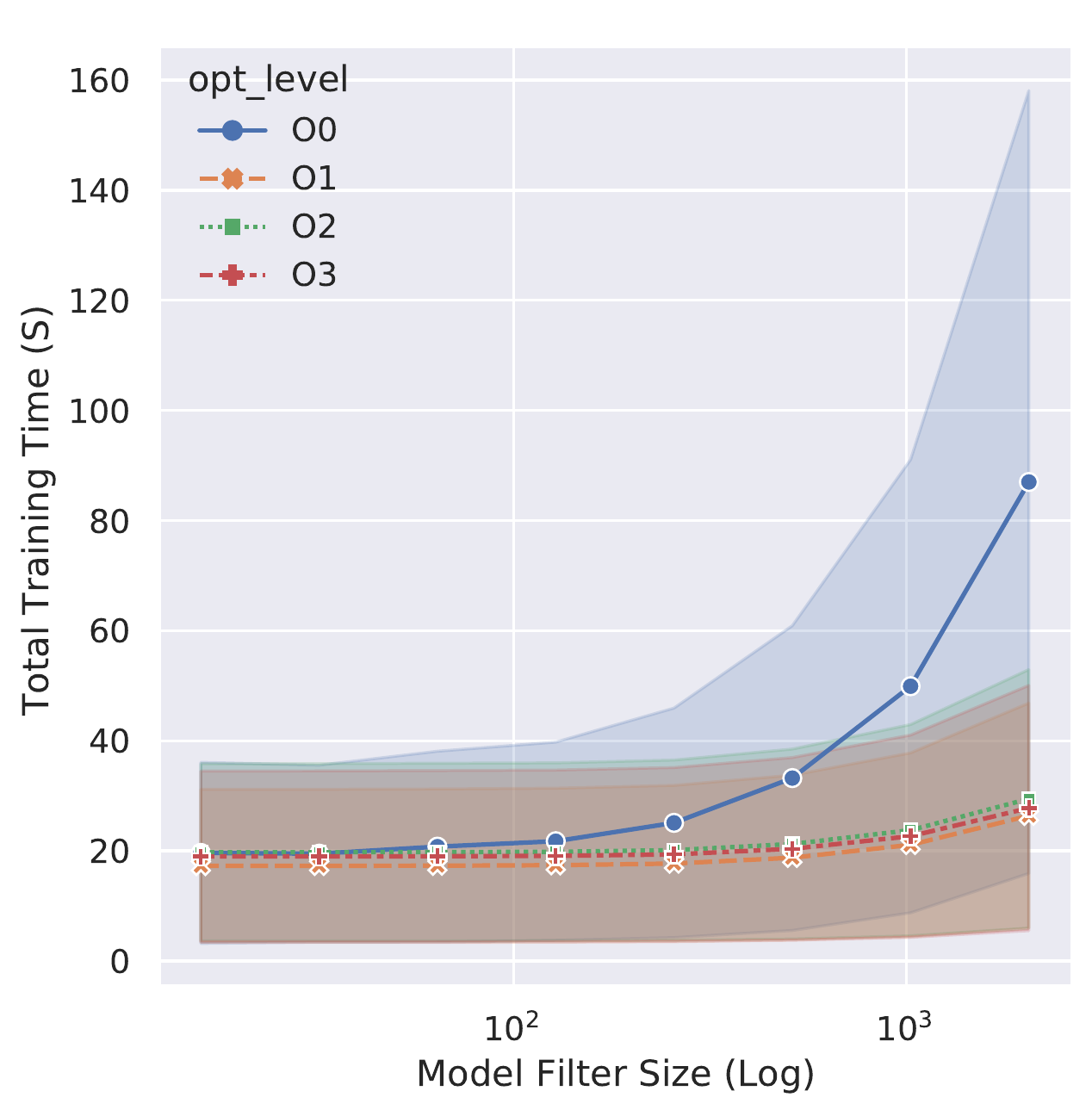}
      \caption{Titan RTX}
    \end{subfigure}
    \begin{subfigure}[b]{0.28\textwidth}
      \includegraphics[width=\textwidth]{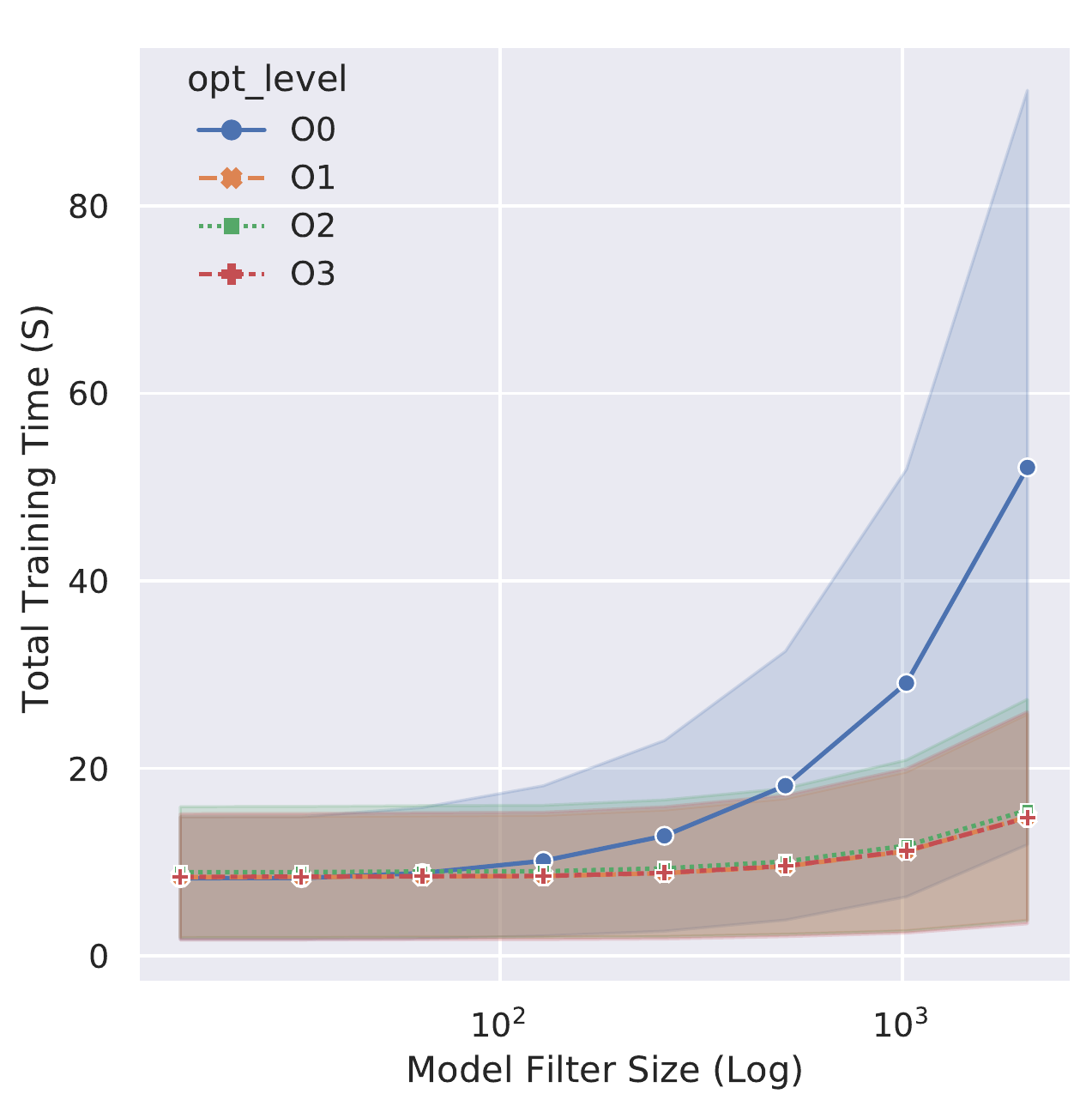}
      \caption{V100}
    \end{subfigure}
    \begin{subfigure}[b]{0.28\textwidth}
      \includegraphics[width=\textwidth]{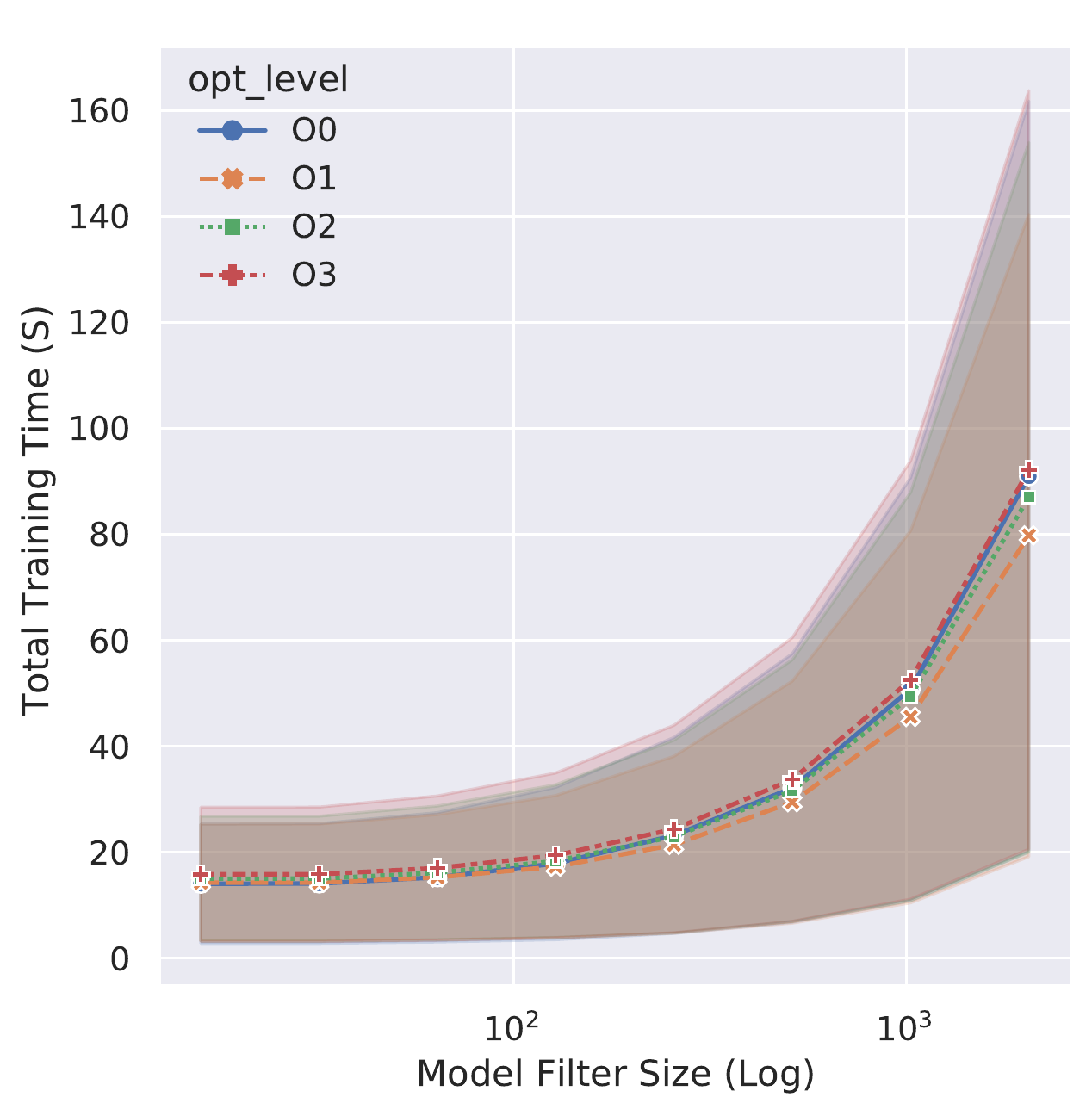}
      \caption{P100}
    \end{subfigure}
    \caption{Total training time as the model size is increased for the GAE model.}
    \label{fig:gae-model-time}
\end{figure*}

\begin{figure*}[h]
  \centering
    \begin{subfigure}[b]{0.28\textwidth}
      \includegraphics[width=\textwidth]{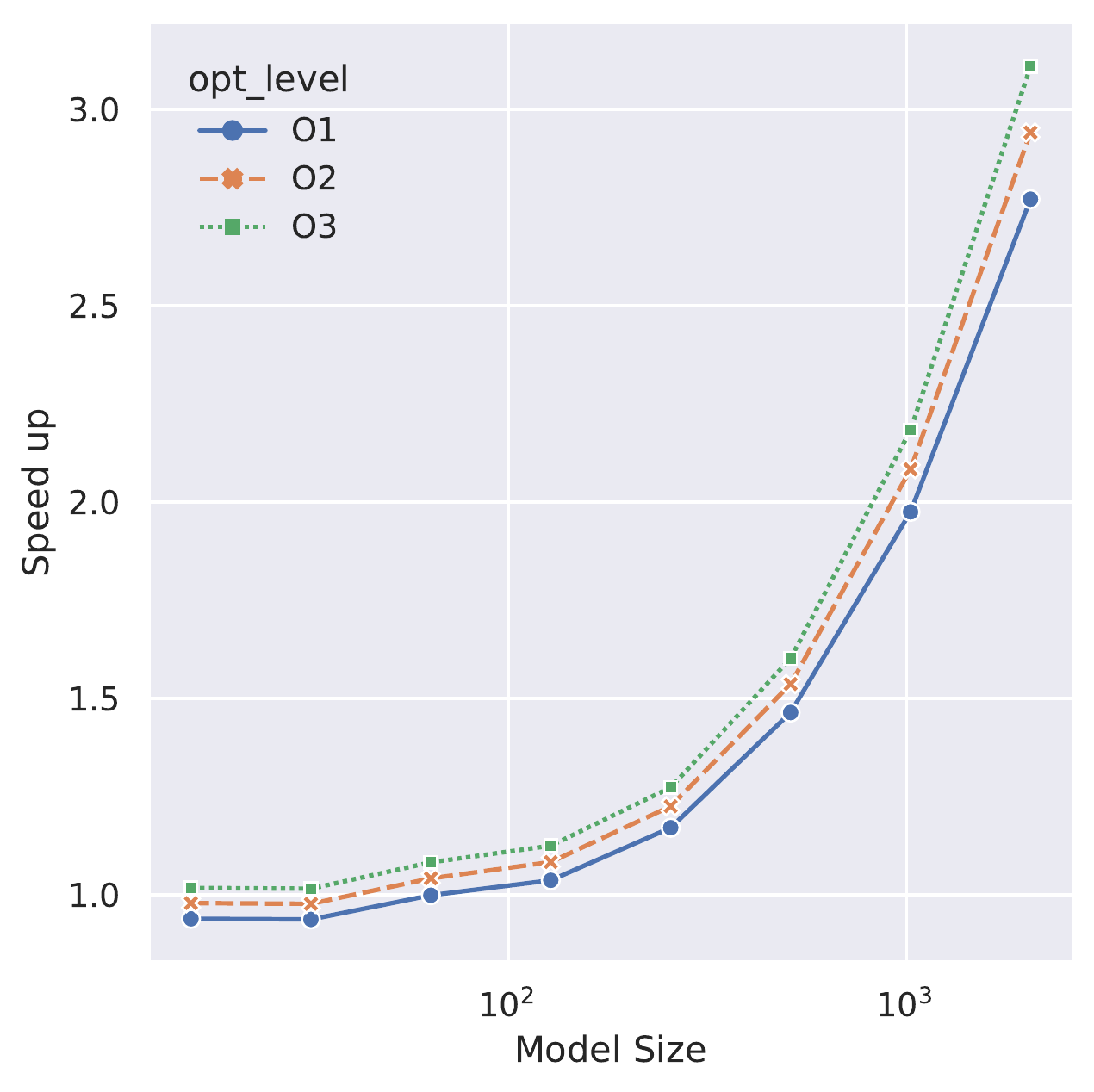}
      \caption{Titan RTX}
    \end{subfigure}
    \begin{subfigure}[b]{0.28\textwidth}
      \includegraphics[width=\textwidth]{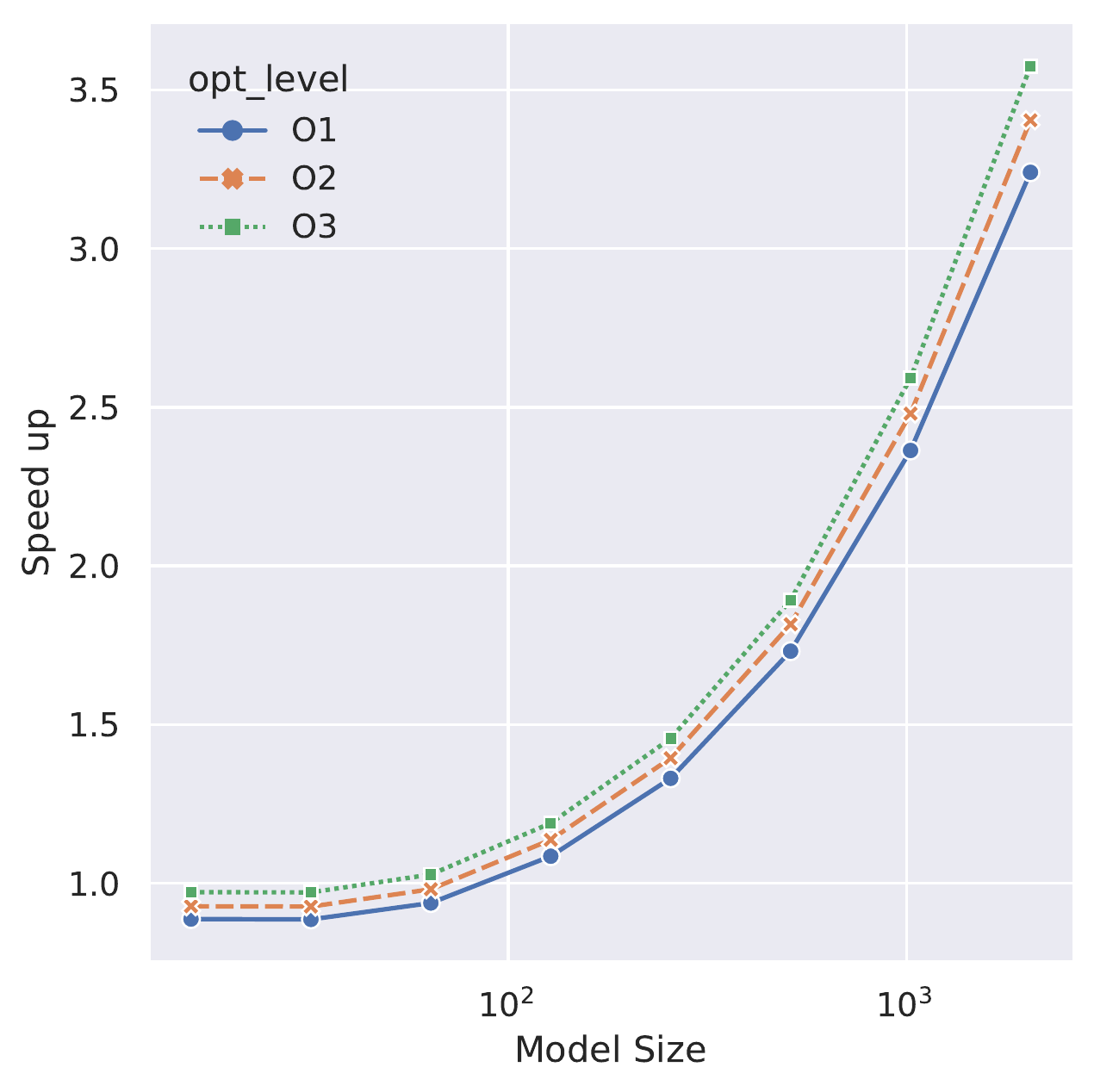}
      \caption{V100}
    \end{subfigure}
    \begin{subfigure}[b]{0.28\textwidth}
      \includegraphics[width=\textwidth]{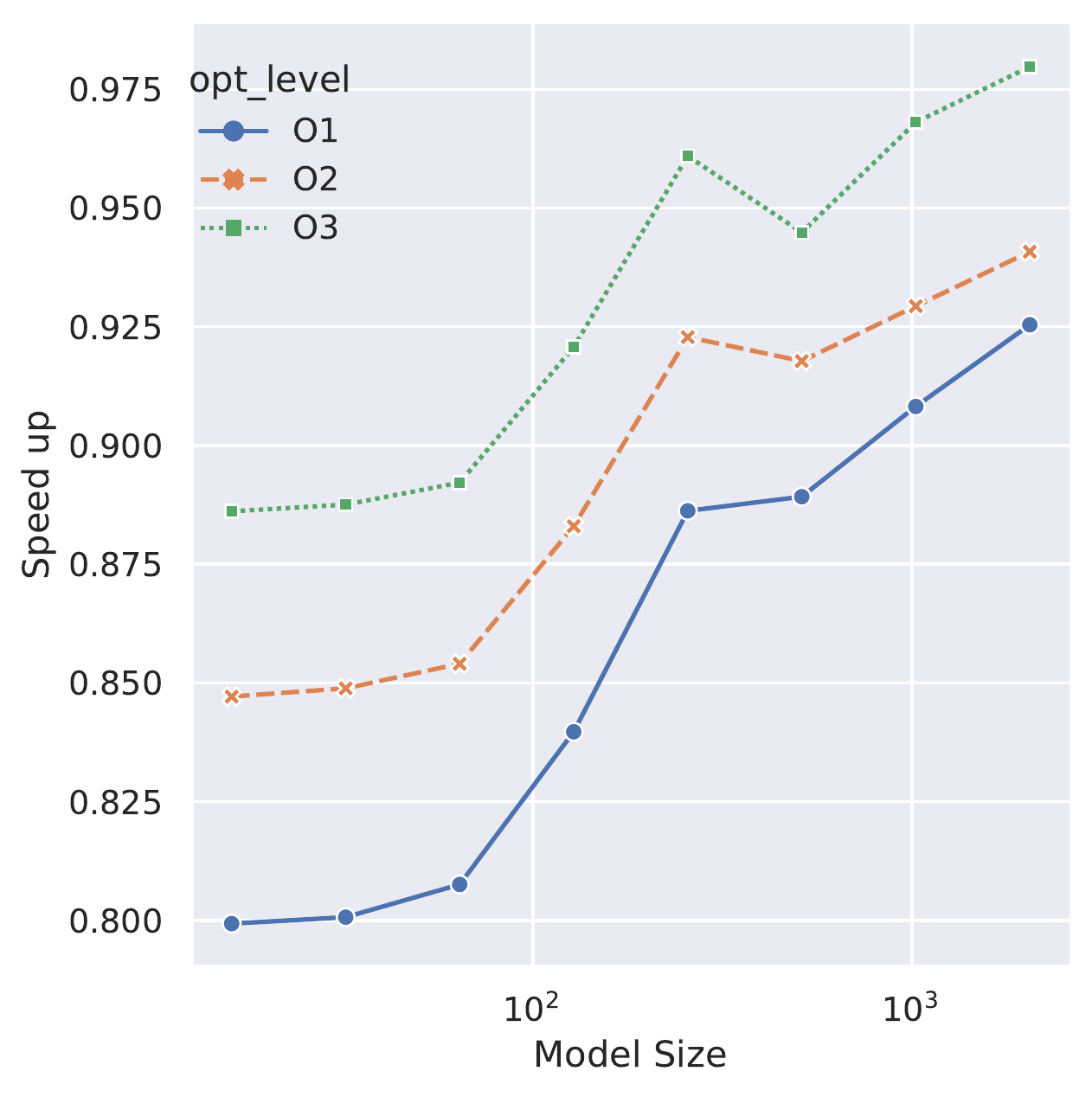}
      \caption{P100}
    \end{subfigure}
    \caption{Speed up of the various opt levels versus O0 for the GAE approach as model size increases.}
    \label{fig:gae-model-speedup}
    \vskip -10pt
\end{figure*}

Figure \ref{fig:gae-model-time} illustrates how various GPUs scale across model sizes in the GAE approach. The overall trend is similar to that of the GCN approach, with the full-precision mode demonstrating a large and sharp increase in run-time as larger model sizes are reached. However, one key difference is the larger variance displayed at each point, meaning the run-time for the GAE approach is more sensitive to the input graph size. Figure \ref{fig:gae-model-speedup} shows the speed up versus the full-precision baseline for all cards. Two interesting trends can be observed from the figure: firstly, the difference in speed up between the various optimisation levels is reduced here versus the GCN results, secondly the speed up is overall less for a given model size than was shown for the GCN result. These results suggest that, due to the complex graph reconstruction method used for the model optimisation, GAE type models are more sensitive to the input graph size than the overall model complexity.

\section{Conclusion}
\label{sec:conclusion}

In this work, we have attempted to provide a clear and detailed analysis of the impact of using reduced-precision computation and specialised GPU hardware designed for such operations on graph-based neural networks. Tensor Cores, introduced in modern NVIDIA GPU architectures, are capable of enabling mixed-precision operations by dynamically adapting calculations to accelerate throughput while preserving accuracy. While the effects of these improvements have been thoroughly explored in various facets of machine learning, such as computer vision and natural language processing, definitive literature on graph convolutional neural networks, which could theoretically benefit from reduced precision and Tensor Cores, is sparse. In this vein, we perform comprehensive experiments to evaluate the effects of using mixed-precision training on the predictive performance of both the semi-supervised classification and link prediction tasks in graph neural networks. We also measure the change in both run-time and the maximum memory consumed on the GPU as the number of vertices in the input graph and the model size is increased for the various levels of reduced-precision optimisation.

As expected, our experiments demonstrate that using reduced-precision optimisation modes, taking advantage of Tensor Cores, reduce run-time for all models training by a significant margin. This points to the great advantage that reduced-precision and Tensor Cores can provide, considering certain layer parameters are divisible by 8. As for memory usage, our experiments indicate an adverse impact of using automatic mixed precision on graph convolutional networks. Using mixed-precision (O1,O2) increases memory usage compared to using full-precision (O0) or half-precision (O3) operations. In terms of the predictive performance of the models, we observe that using complete half-precision (O3) fully hampers the learning process and unsurprisingly leads to a total model collapse, while using mixed-precision (O1,O2) indicates no significant change in performance compared to the full-precision mode (O0).

While this work has been primarily focused on NVIDIA Apex enabling automatic mixed-precision via PyTorch, other frameworks and libraries making use of various other forms of reduced precision need to be fully investigated, which would be an interesting trajectory for future work.

\section*{Acknowledgement}

We gratefully acknowledge the support of the Engineering and Physical Sciences Research Council UK (EPSRC) for funding (EP/M020576/1, 1444756).

\bibliographystyle{IEEEtran}
\bibliography{paper_ref}

\end{document}